\RequirePackage{amsmath,amsthm}
\PassOptionsToPackage{capitalize}{cleveref}
\documentclass{bytedance_seed}

\usepackage{amsmath,amsfonts,bm}

\def\eqref#1{equation~\ref{#1}}

\def\1{\bm{1}}

\DeclareMathAlphabet{\mathsfit}{\encodingdefault}{\sfdefault}{m}{sl}
\SetMathAlphabet{\mathsfit}{bold}{\encodingdefault}{\sfdefault}{bx}{n}

\usepackage{hyperref}
\usepackage{url}
\usepackage{xcolor}         %

\usepackage{url}            %
\usepackage{booktabs}       %
\usepackage{graphicx}
\usepackage{multirow}
\usepackage{amsmath}
\usepackage{amssymb}
\usepackage{bm}
\usepackage{makecell}
\usepackage[ruled, lined, linesnumbered, commentsnumbered, longend]{algorithm2e}
\usepackage{tocloft}
\usepackage{xspace}
\usepackage{amsthm}
\usepackage{wrapfig}
\usepackage{microtype}
\usepackage{aliascnt}
\usepackage{enumitem}
\usepackage[nameinlink,capitalize,noabbrev]{cleveref}

\def\method{DMAD\xspace}

\newcommand{\dmdtwo}{DMD2\xspace}

\newcommand{\ie}{\textit{i.e.}\xspace}

\newaliascnt{proposition}{theorem}
\newtheorem{proposition}[proposition]{Proposition}
\aliascntresetthe{proposition}
\newaliascnt{lemma}{theorem}

\aliascntresetthe{lemma}
\newaliascnt{corollary}{theorem}

\aliascntresetthe{corollary}

\title{DMAD: Distribution Matching as Adversarial Distillation for Fast Visual Generation}

\renewcommand\authorformat[2][]{\authorfont{\sffamily\mbox{#2$^{#1}$}}}
\author[1,2]{Zhengming Yu}
\author[2,\dagger]{Junkun Yuan}
\author[2]{Haotian Yang}
\author[2]{Gordon Guocheng Qian}
\author[2]{Yizhi Wang}
\author[2]{Angtian Wang}
\author[2]{Yiding Yang}
\author[2]{Bo Liu}
\author[1]{Xin Li}
\author[1]{Wenping Wang}
\author[2]{Chongyang Ma}
\affiliation[1]{Texas A\&M University}
\affiliation[2]{Intelligent Creation, ByteDance}
\affiliation[\dagger]{Project lead}

\fancypagestyle{firststyle}{%
  \fancyhead[L]{\includegraphics[width=55.3mm,height=10mm]{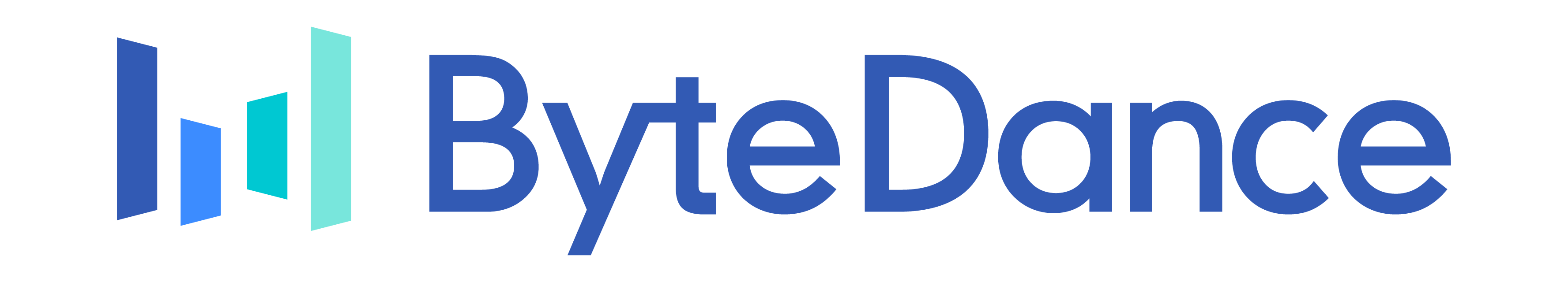}}%
  \fancyhead[R]{\includegraphics[trim=20bp 424bp 20bp 424bp,clip,height=10mm,keepaspectratio]{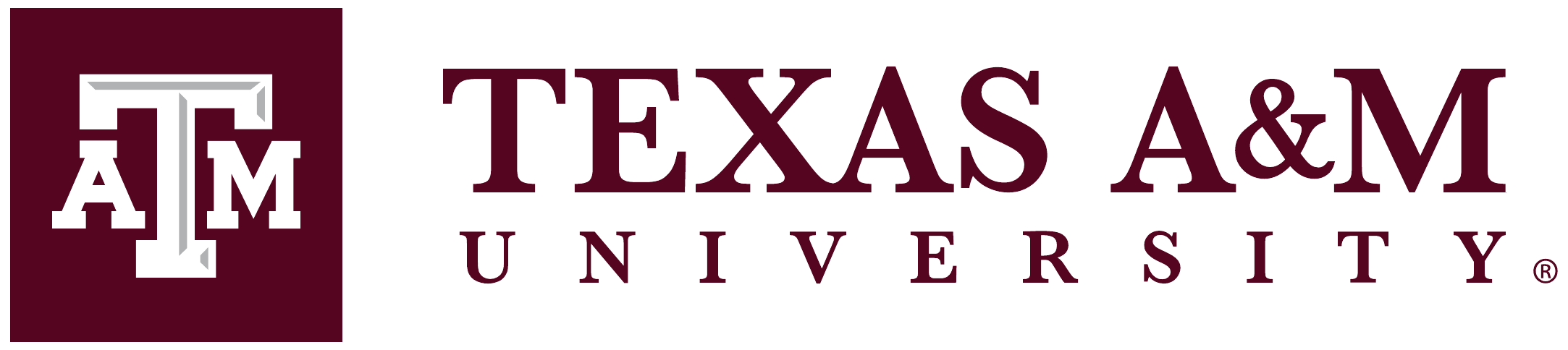}}%
}
\abstract{Distribution Matching Distillation (DMD) trains a few-step student from the difference between separately estimated target and student scores, so it must keep an auxiliary diffusion model fitted to the student's evolving distribution at extra memory and computation cost. We introduce \method, \textbf{D}istribution \textbf{M}atching as \textbf{A}dversarial \textbf{D}istillation, which recasts distribution matching as classification and learns the required log-density ratios directly. Two discriminator heads on a shared backbone distinguish real data and teacher samples from the student's, and linear losses on their logits train the student without auxiliary score fitting. We prove that at the discriminator optimum these losses recover the distribution-matching gradient underlying DMD, through the classical identity linking discriminator logits to log-density ratios. We further introduce gap-based reweighting, which adapts teacher supervision across noise levels from the real-data head's empirical logit gap between real and teacher samples. \method reaches a Fr\'echet Inception Distance (FID) of $1.04$ with one-step generation on ImageNet-$64{\times}64$, $14.47$ with four-step SDXL on COCO-10K, and a VBench total score of $85.15$ with four-step Wan2.1-T2V-14B, the best values among the compared few-step methods and the multi-step teachers. On MiniMax-H3-33B, our four-step student achieves overall human preference rates of $79.1\%$ over \dmdtwo and $84.6\%$ over rCM for joint audio-video generation, excluding ties. Our code, models and demos are available at \url{https://yzmblog.github.io/projects/DMAD}.}
\begin{document}

\maketitle

\begin{figure}[t]
    \centering
    \includegraphics[width=\linewidth]{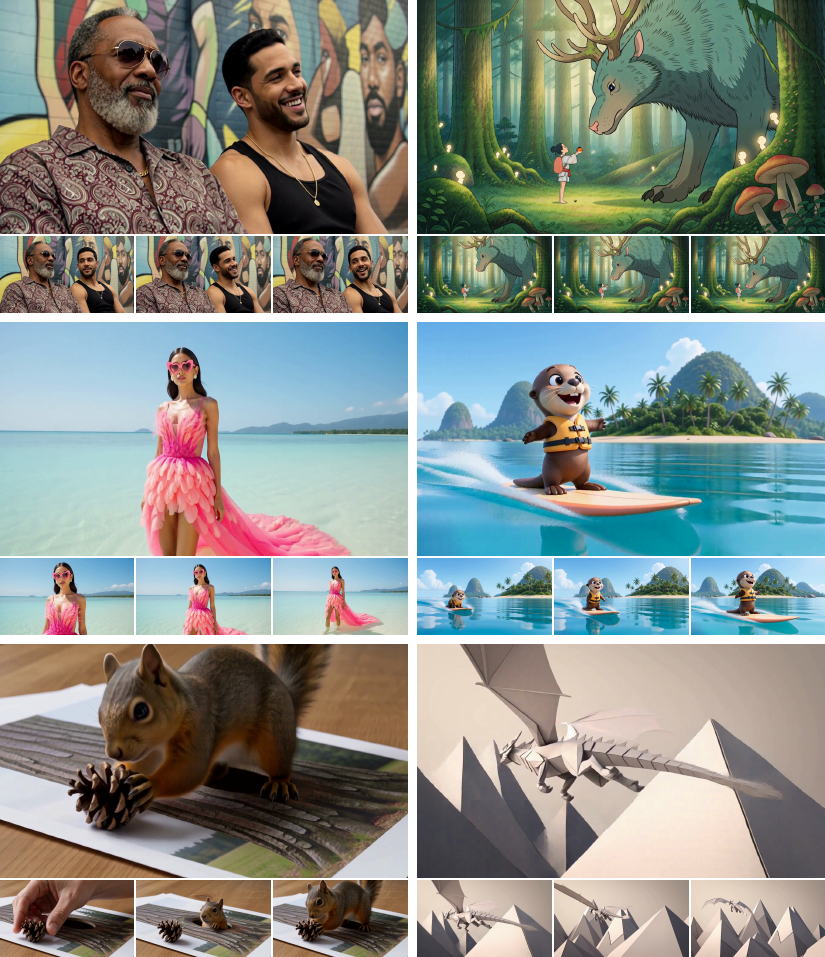}
    \caption{\textbf{Samples of our four-step student distilled from MiniMax-H3.} Zoom in for details.}
    \label{fig:teaser}
\end{figure}

\section{Introduction}
\label{sec:intro}

Diffusion models generate high-quality images and
videos~\citep{ho2020denoising,song2020score,podell2023sdxl,wan2025wan},
but sampling requires many sequential denoising steps.
Step distillation reduces this cost by training a few-step generator from a
pretrained teacher. Existing approaches supervise the student along the
teacher's sampling trajectories~\citep{salimans2022progressive,song2023consistency,kim2023consistency}
or match distributions~\citep{yin2024onestep,yin2024improved}.
Distribution matching lets the student learn its own mapping from noise to
samples while preserving the target distribution.

Distribution Matching Distillation (DMD)~\citep{yin2024onestep} trains the
student on the difference between target and student scores.
A frozen teacher supplies the target score, while an auxiliary diffusion
model, the critic, tracks the evolving student distribution. When the critic
lags behind, its score estimate carries error and training destabilizes. \dmdtwo~\citep{yin2024improved} mitigates this with a two
time-scale update rule (TTUR) that runs five critic updates per student
update. Stability improves, but the student score is still only approximated,
and each student update costs five critic passes. Beyond the critic, \dmdtwo
evaluates the teacher online during training, adding further memory and
computation cost. Its denoising task also constrains the architecture of the
discriminator.

We introduce \method, Distribution Matching as Adversarial Distillation,
which replaces separate score estimation with two classification tasks, one
against teacher samples and one against real data. Two discriminator heads
distinguish these targets from student samples, and linear losses on their
logits drive the updates. At the discriminator optimum, this
loss yields the distribution-matching gradient underlying DMD
(\cref{prop:dmd-gan}). \method eliminates auxiliary score fitting.
In our experiments, we use one discriminator update per student update.
Freed from the denoising task, the discriminator can build on diffusion
encoders or pretrained classification features. The teacher now only supplies samples, so we generate
them once and cache them, removing online teacher evaluation from training.

\begin{wrapfigure}[23]{r}{0.45\textwidth}
    \centering
    \includegraphics[width=\linewidth]{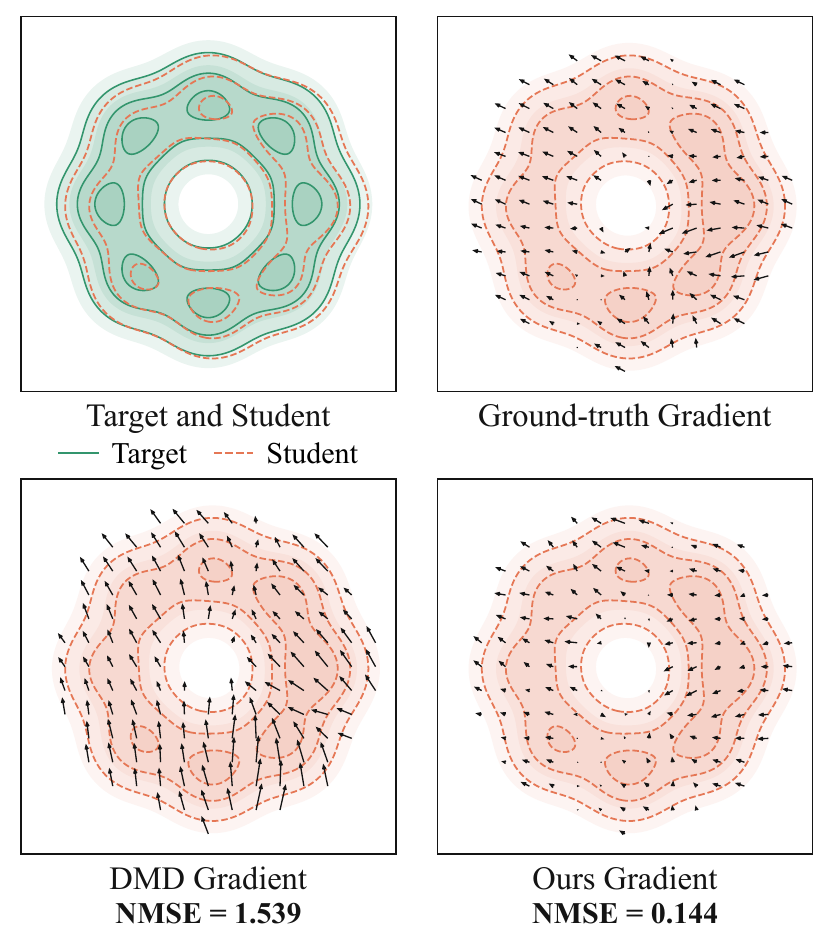}
    \caption{\textbf{Gradient estimation on two Gaussian mixtures with a fixed student.}
    Our logit gradient more closely matches the ground-truth field
    than DMD's score difference.}
    \label{fig:toy-gradient}
\end{wrapfigure}

Errors in separately estimated scores can be large relative to their
difference, which drives the student update.
We examine this in a controlled experiment on two nearby Gaussian mixtures
with known gradients, holding the student distribution fixed and giving
both estimators the same teacher pretraining and seed.
\Cref{fig:toy-gradient} compares DMD's score difference with our single-head
logit gradient. In this setting, our estimate follows the ground-truth field
more closely, with normalized mean squared error of $0.144$, compared with
$1.539$ for DMD.
\Cref{sec:toy-gradient-details} gives the setup and the analytic gradients.

\method draws on two targets, the teacher's samples and real data, which
play complementary roles. Previous methods, including
\dmdtwo~\citep{yin2024improved}, SiDA~\citep{zhou2024adversarial}, and
SenseFlow~\citep{ge2025senseflow}, add adversarial supervision on top of a
score-based objective. In contrast, our discriminator heads carry the
distribution-matching gradient themselves. Each head compares student samples
with one target, and a linear loss on their logits trains the generator. The
real-data head supplies supervision beyond the teacher's sampled distribution.
Teacher samples alone would bound the student by the teacher's own fidelity;
training against real data lifts that bound, and our students outperform their
teachers on both ImageNet and COCO.
Our ablations in \cref{tab:ablation} demonstrate the complementary
roles of the two targets: removing the real-data head raises four-step FID from $14.47$ to $21.37$ on COCO-10K, while removing teacher supervision reduces semantic fidelity and visual quality, dropping VBench total score from $84.70$ to $82.20$ and semantic score from $79.55$ to $75.36$.

The balance between the two targets can vary across noise levels. During training, the real-data head assigns different mean logits to real and teacher samples (\cref{fig:gap_fid}(a)). This gap varies across noise levels and depends on the current student and discriminator. We use it as an empirical signal for \emph{gap-based reweighting}, assigning larger teacher weights to noise bands with lower gaps. This adaptive allocation improves performance for both image and video generation (\cref{tab:ablation}).

Our contributions are:
\begin{itemize}[leftmargin=1.4em,itemsep=1pt,topsep=2pt]
    \item We introduce \method, a distribution-matching distillation
    framework that jointly learns from teacher-generated samples and real
    data through two binary classification tasks, eliminating auxiliary score
    fitting and online teacher evaluation.

    \item We develop gap-based reweighting, which uses the real-data head's
    mean responses to teacher and real samples within each noise band to
    adapt the teacher's supervision as training proceeds.

    \item \method achieves $1.04$ FID with one-step generation on
    ImageNet-$64{\times}64$ and $14.47$ with four-step SDXL on COCO-10K,
    where one step reaches $16.11$.
    Four-step Wan2.1-T2V-1.3B and -14B reach VBench total scores of
    $84.70$ and $85.15$. These are the best values among the compared few-step
    methods and the multi-step teachers. On the 33B
    MiniMax-H3~\citep{minimaxh32026}, human raters prefer our four-step student
    to \dmdtwo in $79.1\%$ and to
    rCM~\citep{zheng2026large} in $84.6\%$, excluding ties.
\end{itemize}

\section{Related Work}
\label{sec:related}

\paragraph{Trajectory and consistency methods.}
Early work distills a teacher either by regressing to its outputs~\citep{luhman2021knowledge,meng2023distillation} or by progressively reducing the number of sampling steps~\citep{salimans2022progressive}. PDD~\citep{shaul2026parallel} extends this idea to video using multi-interval mean velocities. Consistency and flow-map methods impose consistency along a common trajectory, either by mapping different states to the same endpoint or by relating states at different points on that trajectory~\citep{song2023consistency,song2024improved,kim2023consistency}. sCM~\citep{lu2025simplifying} extends consistency modeling to continuous time, while MeanFlow~\citep{geng2025mean} enables one-step generation from scratch. These methods all constrain how the student follows a particular trajectory. Distribution matching instead constrains the distribution of its outputs.

\paragraph{Score-based distribution matching.}
Diff-Instruct~\citep{luo2023diff}, VSD~\citep{wang2023prolificdreamer}, and DMD~\citep{yin2024onestep} optimize a generator through score-based distribution matching. DMD uses the difference between a target score and a score fitted to the student's evolving distribution. \dmdtwo~\citep{yin2024improved} removes the regression loss and strengthens online critic training, while SenseFlow~\citep{ge2025senseflow} improves stability for large flow-based models. SiD~\citep{zhou2024score} and SGMD~\citep{wu2026sgmd} derive alternative objectives from score identities and Fisher-based matching. DP-DMD~\citep{wu2026diversity} and PhasedDMD~\citep{fan2026phased} distribute supervision across generation steps or noise intervals. rCM~\citep{zheng2026large} combines score regularization with consistency training, AnyFlow~\citep{gu2026anyflow} learns flow-map transitions for any-step video generation, and AMD~\citep{bai2026optimizing} adapts distribution matching using reward proxies. These methods still rely on score estimation. FSF-DMD~\citep{kim2026distribution} removes the auxiliary fake-score network but derives a fake-score surrogate from the generator's own flow map. Our method instead estimates no score and learns the required log-density ratio directly through classification.

\paragraph{Adversarial distillation.}
Adversarial distillation typically trains discriminators against real images or videos~\citep{sauer2023adversarial,lin2025diffusion,lin2025autoregressive}, or uses intermediate teacher predictions as adversarial targets~\citep{sauer2024fast,lin2024sdxl}. D2O-F~\citep{zheng2025revisiting} finetunes a pretrained diffusion model with a projected discriminator. \dmdtwo, SiDA~\citep{zhou2024adversarial}, SenseFlow, and ADM~\citep{lu2025adversarial} add adversarial supervision to a score-based distribution-matching objective. DDO~\citep{zheng2025direct} connects likelihood ratios to an implicit discriminator, while $f$-distill~\citep{xu2025one} uses density ratios to reweight score differences. Unlike these methods, we do not use adversarial training to supplement a score-based objective. The discriminator logits themselves provide the distribution-matching gradient, eliminating student-score estimation. Separate heads match teacher samples and real data, with their balance adapted across noise levels.

\section{Method}
\label{sec:method}

\begin{figure}[t]
    \centering
    \includegraphics[width=\linewidth]{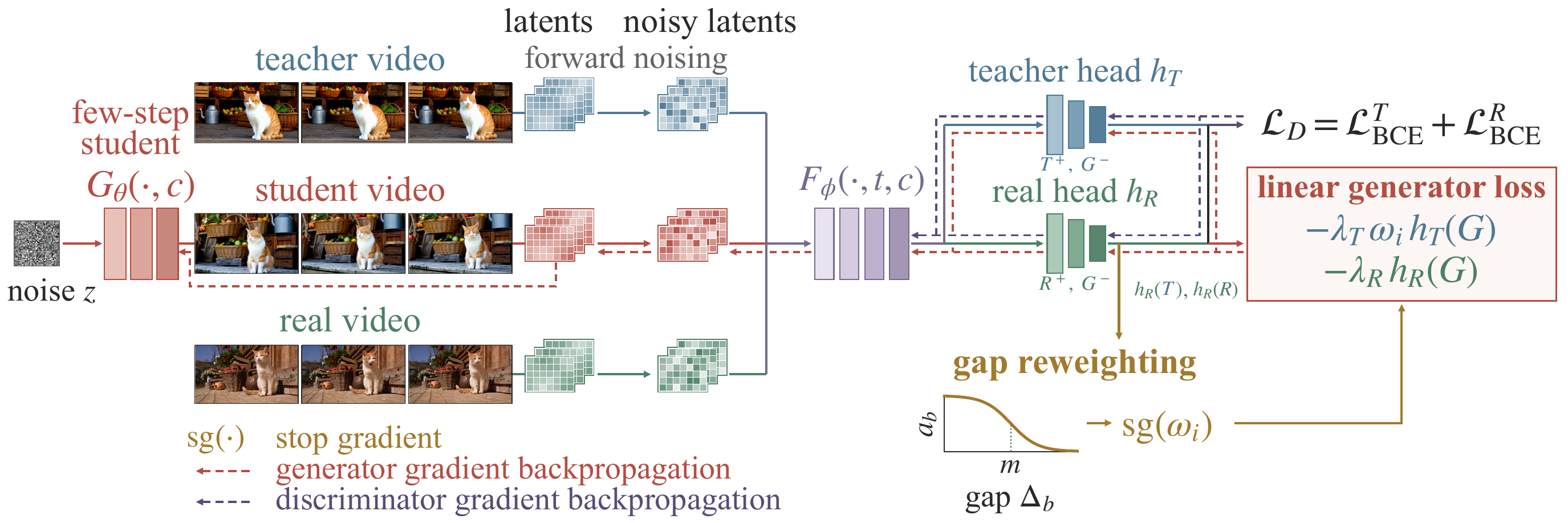}
    \caption{\textbf{Overview of \method (drawn for video distillation).}
    $T$, $R$, and $G$ denote noisy teacher, real, and student samples, with
    time and text conditions omitted. A shared backbone $F_\phi$ feeds a teacher
    head and a real head, and a linear loss on each logit trains the generator.
    The real-data head also provides the gap for teacher reweighting. Video frames are
    decoded only for illustration.}
    \label{fig:method_video}
\end{figure}

We reformulate DMD's score-estimation pipeline as joint classification-based
matching to teacher samples and real data. Linear losses on two discriminator logits train the few-step student. The same real-data head also scores teacher samples, and the gap it leaves between them and real data adapts teacher supervision across noise levels. \Cref{fig:method_video} illustrates the pipeline for video distillation. Image distillation is the same, with only the discriminator heads adapted to the backbone (\cref{sec:impl}).

\subsection{Reformulating Distribution Matching Distillation}
\label{sec:method-dmd}
\label{sec:method-equiv}

We start from the idealized distribution-matching update underlying
DMD~\citep{yin2024onestep}. Let $G_\theta$ be a student with parameters $\theta$ and the final rollout output
$\bm x_0=G_\theta(\bm z,\bm c)$ for initial noise $\bm z$ and condition
$\bm c$, inducing a distribution $p_\theta$. At noise level $t$, we form
$\bm x_t=\alpha_t\bm x_0+\sigma_t\bm\epsilon$, where
$\bm\epsilon\sim\mathcal N(\bm0,\bm I)$ is independent Gaussian noise.
The forward schedule specifies signal scale $\alpha_t$ and noise standard
deviation $\sigma_t$. Let $p_t$ be the noisy marginal of a clean distribution
$p$ and $\bm s_p=\nabla_{\bm x_t}\log p_t$ its score.
All densities are conditional on $\bm c$, omitted below. As in DMD, the
analysis assumes the same fixed distribution over $t,\bm c$ for target and
student samples, and all expectations below use it.

For a fixed target distribution $p$, the update above is motivated by the
reverse-KL objective
\begin{equation}
    \mathcal L_{\mathrm{DM}}(\theta;p)
    =\mathbb E_{t,\bm c}
    \left[D_{\mathrm{KL}}\!\left(p_{\theta,t}\,\middle\|\,p_t\right)\right].
    \label{eq:dmd-objective}
\end{equation}
Its generator gradient depends on the score difference, with
$\bm s_\theta=\nabla_{\bm x_t}\log p_{\theta,t}$ the student score:
\begin{equation}
    \nabla_\theta\mathcal L_{\mathrm{DM}}
    =-\mathbb E_{\bm z,\bm\epsilon,t,\bm c}
    \left[
      \left(\frac{\partial\bm x_t}{\partial\theta}\right)^{\!\mathsf T}
      \big(\bm s_p(\bm x_t,t,\bm c)-\bm s_\theta(\bm x_t,t,\bm c)\big)
    \right].
    \label{eq:kl-grad}
\end{equation}
DMD estimates these two scores separately using a frozen teacher and a critic trained on student samples. \method instead recovers the score difference directly from a discriminator logit.

To estimate the required score difference directly, we train a single binary
classifier to tell target samples from student samples at each noise level.
Let $h_\phi$ be the scalar logit of a discriminator $D$ with parameters
$\phi$, and $D_\phi=S(h_\phi)$ its target-class probability, where
$S(u)=(1+e^{-u})^{-1}$ is the logistic sigmoid function. We
minimize the balanced binary cross-entropy (BCE) $\mathcal B(h_\phi;p)$, which
weights target and student samples equally, so its optimum is the log-density
ratio alone (\cref{sec:proof}).

\begin{proposition}[Classification-based distribution-matching update]
\label{prop:dmd-gan}
At fixed $t,\bm c$, assume positive differentiable densities on common
support and the regularity needed to exchange differentiation and
integration. The unrestricted, unregularized BCE optimum is
$h^\star=\log p_t-\log p_{\theta,t}$, so
\begin{equation}
    \nabla_{\bm x_t}h^\star
    =\bm s_p-\bm s_\theta.
    \label{eq:logit-score}
\end{equation}
At the current parameters, the generator gradients of
$-\mathbb E[h^\star]$ and $\mathcal L_{\mathrm{DM}}$ are equal with
$h^\star$ fixed.
\end{proposition}

The proof follows from the optimal discriminator
identity~\citep{goodfellow2014generative} and the chain rule, and is given in
\cref{sec:proof}. We therefore update the generator with the linear loss
\begin{equation}
    \mathcal L_G^{\mathrm{lin}}(\theta;\phi)
    =-\mathbb E_{\bm z,\bm\epsilon,t,\bm c}
      [h_\phi(\bm x_t,t,\bm c)].
    \label{eq:linear-loss}
\end{equation}
With $\phi$ fixed, gradients through the discriminator's input train the
generator without auxiliary score fitting, and the learned discriminator
approximates the ideal logit gradient (\cref{fig:toy-gradient}). Unlike the
linear loss, the non-saturating loss $-\log D_\phi$ multiplies each sample's
logit-gradient contribution by $1-D_\phi$. We analyze this weighting in
\cref{sec:proof} and compare the losses in \cref{sec:ablation}. One classifier
therefore replaces both score estimates, and the linear loss carries its
gradient unweighted.

\subsection{Distillation with Teacher and Real Targets}
\label{sec:method-two-targets}

To provide supervision beyond the teacher's sampled distribution, we
train on both teacher-generated samples and real data, with distributions
$p_{T}$ and $p_{R}$.
A shared backbone $F_\phi$ feeds two heads with scalar logits
$h_{T}$ and $h_{R}$, each trained to distinguish its own
target samples from student samples. The combined discriminator objective is
the sum of the two balanced classification losses:
\begin{equation}
    \mathcal L_D
    =\mathcal L_{\mathrm{BCE}}^{T}
     +\mathcal L_{\mathrm{BCE}}^{R},
    \label{eq:dual-disc-loss}
\end{equation}
where $\mathcal L_{\mathrm{BCE}}^{T}=\mathcal B(h_{T};p_{T})$
and $\mathcal L_{\mathrm{BCE}}^{R}=\mathcal B(h_{R};p_{R})$.
With fixed nonnegative coefficients
$\lambda_{T}$ and $\lambda_{R}$ setting the relative strength
of the two targets, the generator loss is the linear combination:
\begin{equation}
    \mathcal L_G
    =-\mathbb E\left[
       \lambda_{T}h_{T}(\bm x_t,t,\bm c)
       +\lambda_{R}h_{R}(\bm x_t,t,\bm c)
    \right].
    \label{eq:generator-loss}
\end{equation}
At their optima, each head supplies its own target's score difference
(\cref{eq:logit-score}), so both targets contribute distribution-matching
gradients within one objective.
Our ablations in \cref{tab:ablation} show that real supervision improves
image FID, while teacher supervision preserves detail and text alignment.

\subsection{Gap-Based Reweighting}
\label{sec:method-gap}

In \cref{eq:generator-loss}, $\lambda_{T}$ weights the teacher
equally at every noise level. The real-data head's mean-logit gap between real
and teacher samples varies across noise levels and over training
(\cref{fig:gap_fid}(a)). It depends on the current student and discriminator.
We use this empirical signal to assign larger teacher weights to noise bands
with lower gaps.

To measure the gap at a usable resolution, we group noise levels into bands and
write $b(t)$ for the band containing $t$. For each band $b$, we maintain an
exponential moving average (EMA):
\begin{equation}
    \Delta_b\leftarrow\operatorname{EMA}_{\beta}\!\left[
      \overline{h_{R}(\bm x_t^{R},t,\bm c)}_b
      -\overline{h_{R}(\bm x_t^{T},t,\bm c)}_b
    \right].
    \label{eq:band-gap}
\end{equation}
The bars average the same real head's logits over noisy real samples
$\bm x_t^{R}$ and teacher samples $\bm x_t^{T}$ within a
band. We use EMA decay $\beta=0.99$ and update only bands observed in both
sources.

The gap then sets a raw teacher weight $a_b$, centered at the median gap over
the measured bands $\mathcal R$:
\begin{equation}
    m=\operatorname{median}_{b\in\mathcal R}\Delta_b,
    \qquad
    a_b=S\!\left(\frac{m-\Delta_b}{\tau}\right).
    \label{eq:gap-reweight}
\end{equation}
A band with a lower mean-logit gap receives a larger weight. The temperature
$\tau>0$ sets how sharply the
weight varies around the median, and centering at the median gap keeps the
weights
relative as the gaps drift over the course of training.

We normalize these weights to control their overall scale. For
sample $i$ in a minibatch of size $B$, let $\omega_i=a_{b(t_i)}/C$. For ImageNet
and SDXL, $C=\max(B^{-1}\sum_j a_{b(t_j)},B^{-1})$; for Wan,
$C=\max(|\mathcal R|^{-1}\sum_{b\in\mathcal R}a_b,10^{-3})$.
MiniMax-H3 uses the same band-based normalization with a floor of $10^{-4}$.
Collecting these terms, the final generator loss is
\begin{equation}
    \widehat{\mathcal L}_G=-\frac1B\sum_{i=1}^B
    \left[
      \lambda_{T}\omega_i
        h_{T}(\bm x_{t_i,i},t_i,\bm c_i)
      +\lambda_{R}
        h_{R}(\bm x_{t_i,i},t_i,\bm c_i)
    \right].
    \label{eq:routed-loss}
\end{equation}
where $\bm x_{t_i,i}$ is the noisy student output for sample $i$. We hold the gap
statistics and weights fixed during generator backpropagation, so each
$\omega_i$ scales its term without adding a gradient.

\begin{figure}[t]
    \setlength{\abovecaptionskip}{4pt}
    \setlength{\belowcaptionskip}{5pt}
    \centering
    \includegraphics[width=\linewidth]{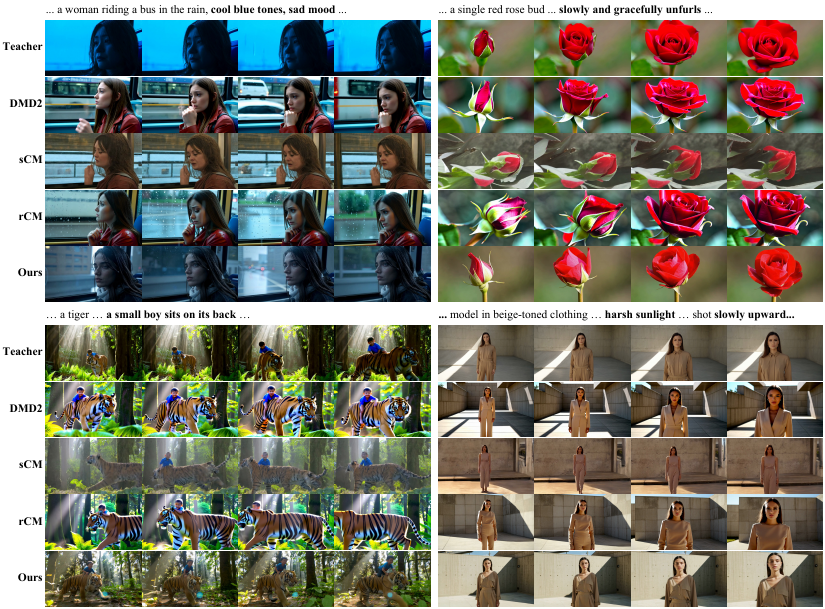}
    \caption{\textbf{Qualitative comparisons of our \method and baseline models on
    Wan2.1-T2V-1.3B.} Rows show the teacher, \dmdtwo, sCM, rCM, and ours. Within
    each example, columns show frames 1, 28, 55, and 81 in order. All compared
    videos are generated using identical initial noise and text prompts. Prompts are abbreviated for presentation.}
    \label{fig:video_comparison}
\end{figure}

\begin{figure}[t]
    \setlength{\abovecaptionskip}{4pt}
    \setlength{\belowcaptionskip}{5pt}
    \centering
    \includegraphics[width=\linewidth]{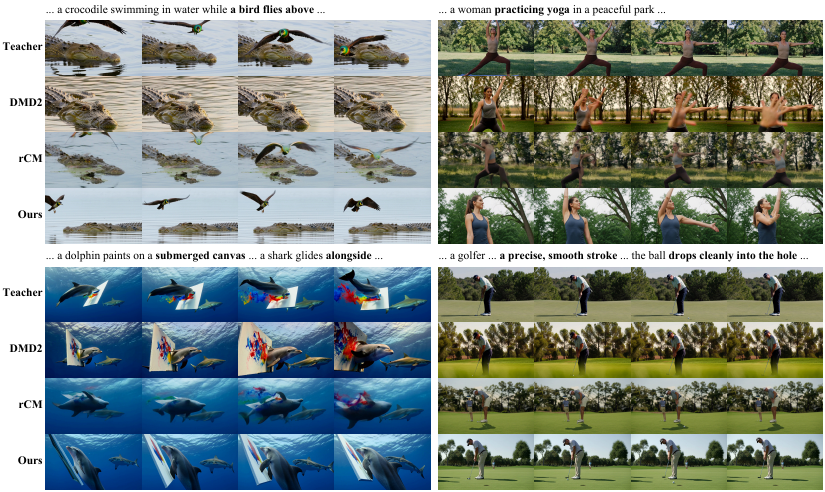}
    \caption{\textbf{Qualitative comparisons of our \method and baseline models on
    MiniMax-H3.} Rows show the teacher, \dmdtwo, rCM, and ours, in that order.
    Within each example, columns show frames 1, 42, 83, and 124 using the same
    indices for all methods. All compared
    videos are generated using identical initial noise and text prompts. Prompts are
    abbreviated for presentation.}
    \label{fig:h3_selected_comparison}
\end{figure}

\section{Experiments}
\label{sec:exp}

\subsection{Experimental Setup}
\label{sec:setup}

We evaluate \method on class-conditional image generation, text-to-image
synthesis, and text-to-video synthesis. Following
\dmdtwo~\citep{yin2024improved}, we distill
EDM~\citep{karras2022elucidating} for
ImageNet-$64{\times}64$~\citep{deng2009imagenet} and
SDXL~\citep{podell2023sdxl} for zero-shot generation on 10K COCO 2014
prompts~\citep{lin2014microsoft}. We use Fr\'echet Inception Distance
(FID)~\citep{heusel2017gans} to measure image quality and diversity, and
CLIP score~\citep{radford2021learning} to evaluate text--image alignment.
For SDXL, we additionally report patch FID~\citep{lin2024sdxl,chai2022any} to
assess high-resolution details, following \dmdtwo. For video, we distill
Wan2.1-T2V 1.3B and 14B~\citep{wan2025wan} to four steps and evaluate them with
VBench~\citep{huang2024vbench} on the same augmented prompt set as
rCM~\citep{zheng2026large}. We further distill MiniMax-H3~\citep{minimaxh32026} to four steps and evaluate them in a user study and AVGen-Bench~\citep{zhou2026avgenbench}. Training and evaluation details are provided in
\cref{sec:impl}. More results are shown in our \textbf {supplementary demo video}.

\subsection{Main Results}
\label{sec:main-results}

\paragraph{Class-conditional image generation.}
\label{sec:imagenet-results}

\Cref{table:imagenet} compares \method with recent baselines.
Our one-step model achieves an FID of $1.24$, outperforming
\dmdtwo, including its longer-training variant, and SiDA trained
directly from the teacher. It also outperforms the EDM teacher with
both ODE and SDE samplers.
With a projected discriminator following D2O-F~\citep{zheng2025revisiting},
\method achieves an FID of $\mathbf{1.04}$, surpassing D2O-F and
SiD$^2$A, which is initialized from a SiD-distilled generator.
The advantage comes from our formulation: the discriminator itself
provides the distillation gradient, so discriminative
representations directly guide generator training. \dmdtwo, in contrast,
requires a critic that must be refitted continually to the student's
evolving distribution, five times per student update.

\begin{table}[t]
\setlength{\abovecaptionskip}{0pt}
\setlength{\belowcaptionskip}{4pt}
\centering
\small
\begin{minipage}[t]{0.48\linewidth}
    \caption{
    \label{table:imagenet}
    \textbf{Comparisons with the EDM~\citep{karras2022elucidating} on ImageNet-64$\times$64.}
    }
    \setlength{\tabcolsep}{4pt}
    \renewcommand{\arraystretch}{1.01}
    \scalebox{0.83}{
    \begin{tabular}{>{\raggedright\arraybackslash}p{2.20in}rr}
    \toprule
    \multirow{2}{*}{Method} & \multirow{2}{*}{\shortstack[c]{\# Fwd \\ Pass~($\downarrow$)}} & \multirow{2}{*}{\shortstack[c]{FID \\ ($\downarrow$)}} \\
    \\
    \midrule
    BigGAN-deep
    \citep{brock2018large}
    & 1 & 4.06 \\
    RIN~\citep{jabri2022scalable} & 1000 & 1.23 \\
    StyleGAN-XL~\citep{sauer2022stylegan} & 1 & 1.52 \\
    \midrule
    Consistency Model~\citep{song2023consistency} & 1 & 6.20\\
    iCT-deep~\citep{song2024improved} & 1 & 3.25 \\
    CTM~\citep{kim2023consistency} & 1 & 1.92 \\
    DMD~\citep{yin2024onestep} & 1 & 2.62 \\
    D2O-F~\citep{zheng2025revisiting} & 1 & 1.16 \\
    DMD2~\citep{yin2024improved} & 1 & 1.51 \\
    DMD2+longer training~\citep{yin2024improved} & 1 & 1.28 \\
    SiD~\citep{zhou2024score} & 1 & 1.52 \\
    SiDA~\citep{zhou2024adversarial} & 1 & 1.35 \\
    SiD$^2$A~\citep{zhou2024adversarial} & 1 & 1.11 \\
    \textbf{\method~(Ours)} & 1 & 1.24\\
     \textbf{+ projected discriminator~(Ours)} & 1 & \textbf{1.04}\\
    \midrule
    EDM~(Teacher, ODE)
    \citep{karras2022elucidating}
    & 511 & 2.32  \\
    EDM~(Teacher, SDE)
    \citep{karras2022elucidating}
    & 511 & 1.36  \\
    \bottomrule
    \end{tabular}}
\end{minipage}
\hfill
\begin{minipage}[t]{0.50\linewidth}
    \centering
    \small
    \caption{
    \label{table:sdxl}
    {\textbf{Comparisons with the SDXL~\citep{podell2023sdxl} on 10K prompts from COCO 2014.}}
    }
    \renewcommand{\arraystretch}{1.15}
    \scalebox{0.83}{
    \begin{tabular}{@{}l@{\hspace{3pt}}r@{\hspace{3pt}}r@{\hspace{3pt}}r@{\hspace{3pt}}r@{}}
    \toprule
    \multirow{2}{*}{Method} & \multirow{2}{*}{\shortstack[c]{\# Fwd \\ Pass~($\downarrow$)}} & \multirow{2}{*}{\shortstack[c]{FID \\ ($\downarrow$)}}  &
    \multirow{2}{*}{\shortstack[c]{Patch \\ FID~($\downarrow$)}}
    &
    \multirow{2}{*}{\shortstack[c]{CLIP \\ ($\uparrow$)}}
    \\
    \\
    \midrule
    \multirow{2}{*}{LCM-SDXL~\citep{luo2023latentlora}} & 1 & 81.62  & 154.40 & 0.275 \\
    & 4 & 22.16  & 33.92 & 0.317 \\
    \midrule
    \multirow{2}{*}{SDXL-Turbo~\citep{sauer2023adversarial}} & 1 & 24.57 & 23.94 & 0.337 \\
    & 4 & 23.19 & 23.27   & 0.334  \\
    \midrule
    \multirow{2}{*}{\shortstack[l]{SDXL\\Lightning~\citep{lin2024sdxl}}}
    & 1 & 23.92  & 31.65 & 0.316 \\
    & 4 & 24.46  & 24.56 & 0.323 \\
    \midrule
    \multirow{2}{*}{\dmdtwo~\citep{yin2024improved}} & 1 & 19.01 & 26.98 & 0.336 \\
    & 4 & 19.32 & 20.86 & 0.332 \\
    \midrule
    SenseFlow~\citep{ge2025senseflow}
    & 4 & 22.83 & 26.05 & 0.335 \\
    \midrule
    \multirow{2}{*}{\textbf{\method~(Ours)}} & 1 & 16.11 & 32.42 & \textbf{0.338} \\
    & 4 & \textbf{14.47} & \textbf{19.88} & 0.328 \\
    \midrule
    SDXL Teacher, cfg=6  & 100 & 19.36 & 21.38 & 0.332 \\
    SDXL Teacher, cfg=8
    & 100  & 20.39 & 23.21 & 0.335 \\
    \bottomrule
    \end{tabular}}
\end{minipage}
\par\vspace{2pt}
\end{table}

\paragraph{Text-to-image synthesis.}
\label{sec:t2i-results}

We follow \dmdtwo~\citep{yin2024improved} for both training data and evaluation.
Our generators are trained by distilling SDXL~\citep{podell2023sdxl}
using 500K real images from LAION-Aesthetics~\citep{schuhmann2022laion} and
corresponding teacher-generated samples, and evaluate on zero-shot COCO
2014~\citep{lin2014microsoft}.
Our four-step generator achieves an FID of $14.47$
and a patch FID of $19.88$ (\cref{table:sdxl}), outperforming \dmdtwo,
SenseFlow~\citep{ge2025senseflow}, and the SDXL teacher on both metrics.
With a single forward pass, \method achieves an FID
of $16.11$ and a CLIP score of $0.338$, the best one-step FID and
highest CLIP score among the compared models.
Qualitative comparisons appear in \cref{fig:sdxl_comparison}.

\begin{figure}[t]
    \setlength{\abovecaptionskip}{2pt}
    \setlength{\belowcaptionskip}{5pt}
    \centering
    \includegraphics[width=\linewidth]{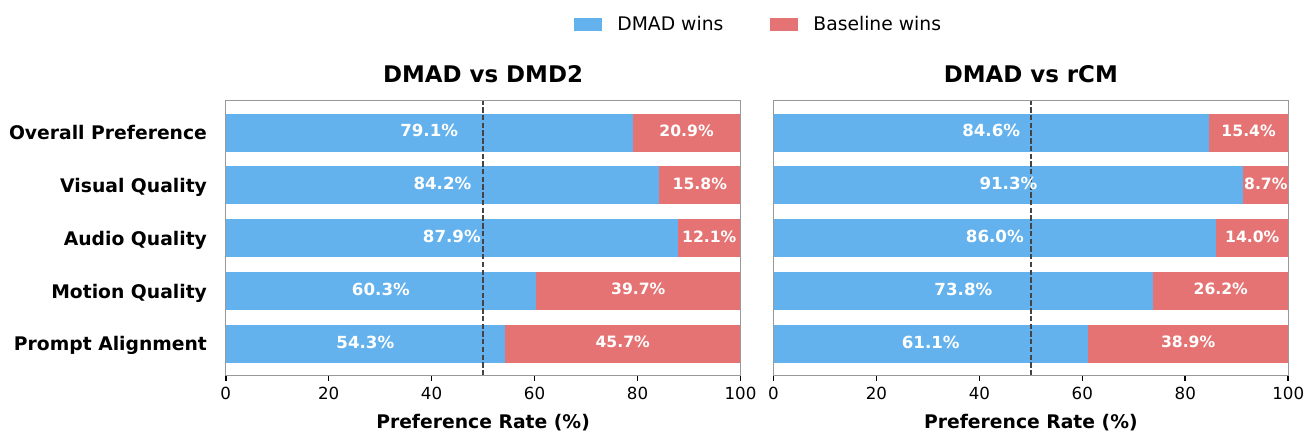}
    \caption{\textbf{User study comparing \method with \dmdtwo and rCM on MiniMax-H3.} All models use 4 sampling steps. Preference rates use prompt-level majority votes, excluding ties. Dashed lines indicate $50\%$.}
    \label{fig:h3_user_study}
\end{figure}

\paragraph{Text-to-video synthesis.}
\label{sec:t2v-results}

We distill Wan2.1-T2V 1.3B, 14B~\citep{wan2025wan} and MiniMax-H3~\citep{minimaxh32026} into four-step
generators using real videos from UltraVideo~\citep{xue2025ultravideo}
and corresponding videos generated by the teacher.
For VBench~\citep{huang2024vbench} evaluation, we generate videos using
the same augmented prompt set as rCM~\citep{zheng2026large}. Our 1.3B model achieves a VBench total score of $84.70$, outperforming
\dmdtwo, sCM~\citep{lu2025simplifying}, rCM, and the 50-step teacher
(\cref{tab:vbench}). At 14B, our model reaches $85.15$, also surpassing
rCM and the teacher. Both models achieve the highest quality scores
among the compared methods.
Additional samples from our four-step Wan2.1-T2V-14B student appear in
\cref{fig:wan14b-samples}.
Qualitative comparisons in \cref{fig:video_comparison} show
more natural colors and better preservation of prompted details,
including petal texture and tiger anatomy.
Our models achieve the highest VBench total and quality scores
among the compared methods, while rCM scores higher on semantics.

For MiniMax-H3, \cref{fig:h3_selected_comparison} shows
improved visual quality and preservation of prompted elements,
such as the bird in the crocodile scene.
A user study on VideoGenEval~\citep{yang2025videogen} prompts
evaluates five dimensions (\cref{fig:h3_user_study}).
\method receives overall preference rates of $79.1\%$ against
\dmdtwo and $84.6\%$ against rCM, excluding ties, with the strongest preferences
in visual and audio quality.
Additional quantitative audio-video results are provided in
\cref{sec:avgenbench}.

\begin{table}[t]
\setlength{\abovecaptionskip}{0pt}
\setlength{\belowcaptionskip}{4pt}
\centering
\caption{\textbf{VBench results for Wan2.1 T2V (480p).}
$^\dagger$Retested with Diffusers and augmented prompts.}
\label{tab:vbench}
\setlength{\tabcolsep}{7pt}
\renewcommand{\arraystretch}{1.15}
\small
\resizebox{\linewidth}{!}{
\begin{tabular}{lcc|c|cc}
\toprule
Model
& Resolution
& NFE
& \makecell[c]{Total~($\uparrow$)}
& \makecell[c]{Quality~($\uparrow$)}
& \makecell[c]{Semantic~($\uparrow$)} \\
\midrule
\multicolumn{6}{c}{Wan2.1 T2V - 1.3B} \\
\midrule
Wan2.1 T2V Teacher~\citep{wan2025wan}$^\dagger$
& $832\times480\times81$ & $50\times2$
& 83.02 & 83.95 & 79.26 \\
DMD2~\citep{yin2024improved}
& $832\times480\times81$ & 4
& 84.56 & 85.58 & 80.50 \\
sCM~\citep{lu2025simplifying}
& $832\times480\times81$ & 4
& 82.06 & 84.00 & 74.30 \\
rCM~\citep{zheng2026large}
& $832\times480\times81$ & 4
& 84.43 & 85.38 & 80.63 \\
\textbf{\method (Ours)}
& $832\times480\times81$ & 4
& \textbf{84.70} & 85.98 & 79.55 \\
\midrule
\multicolumn{6}{c}{Wan2.1 T2V - 14B} \\
\midrule
Wan2.1 T2V Teacher~\citep{wan2025wan}$^\dagger$
& $832\times480\times81$ & $50\times2$
& 83.59 & 84.26 & 80.92 \\
rCM~\citep{zheng2026large}
& $832\times480\times81$ & 4
& 84.92 & 85.43 & 82.88 \\
\textbf{\method (Ours)}
& $832\times480\times81$ & 4
& \textbf{85.15} & 85.96 & 81.90 \\
\bottomrule
\end{tabular}}
\par\vspace{2pt}
\end{table}

\subsection{Ablation Studies}
\label{sec:ablation}

\noindent\begin{minipage}{\linewidth}
We study four ablations in \cref{tab:ablation}: removing the teacher head,
removing the real head, disabling gap reweighting with fixed loss weights,
and replacing the linear generator loss with the non-saturating GAN loss.
Removing a head also removes its discriminator and generator losses. For removing the real head, we also disable gap-based reweighting,
which depends on the real-data head.
\end{minipage}

\begin{table}[t]
\setlength{\abovecaptionskip}{0pt}
\setlength{\belowcaptionskip}{6pt}
\centering
\small
\caption{\label{tab:ablation}\textbf{Ablation experiments of our \method.} One-step image
generation on ImageNet, four-step image generation on COCO-10K (SDXL), and
four-step video generation on VBench (Wan2.1-1.3B).}
\setlength{\tabcolsep}{6pt}
\resizebox{\linewidth}{!}{
\begin{tabular}{lc|ccc|c|cc}
\toprule
& ImageNet & \multicolumn{3}{c|}{COCO-10K (SDXL)} & \multicolumn{3}{c}{VBench (Wan2.1-1.3B)} \\
Setting & FID~($\downarrow$) & FID~($\downarrow$) & Patch FID~($\downarrow$) & CLIP~($\uparrow$)
& Total~($\uparrow$) & Quality~($\uparrow$) & Semantic~($\uparrow$) \\
\midrule
w/o Teacher        & 1.30 & \textbf{13.85} & 26.03 & 0.312 & 82.20 & 83.91 & 75.36 \\
w/o Real           & 1.67 & 21.37 & 23.70 & 0.325 & 84.38 & 85.57 & 79.59 \\
w/o Gap-reweight   & 1.33 & 16.62 & 20.57 & 0.327 & 83.56 & 84.53 & 79.69 \\
w/o Linear loss    & 1.31 & 14.78 & 21.04 & 0.326 & 84.12 & 85.19 & 79.86 \\
\textbf{\method~(Ours)} & \textbf{1.24} & 14.47 & \textbf{19.88} & \textbf{0.328}
& \textbf{84.70} & 85.98 & 79.55 \\
\bottomrule
\end{tabular}}
\par\vspace{12pt}
\end{table}

\begin{figure}[t]
    \setlength{\abovecaptionskip}{6pt}
    \setlength{\belowcaptionskip}{14.3pt}
    \centering
    \includegraphics[width=\linewidth]{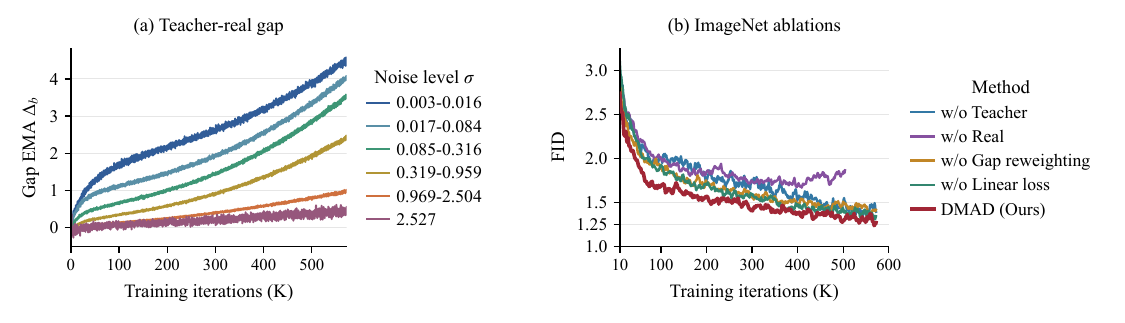}
    \caption{\textbf{ImageNet training dynamics.}
    (a) EMA of the real-head gap between real and teacher samples, the signal
    that drives gap-based reweighting (\cref{sec:method-gap}).
    (b) FID of \method and four ablations without smoothing on a
    1K-iteration grid. The no-real curve ends at 504K iterations.}
    \label{fig:gap_fid}
\end{figure}

Removing the real head causes the largest FID increase on ImageNet and SDXL
in \cref{tab:ablation}. The teacher-only curve in
\cref{fig:gap_fid}(b) also stays above the full model and rises
again after about 400K iterations. Removing the teacher instead slows the
decrease in ImageNet FID. Although it gives the lowest SDXL FID, it also
gives the worst patch FID and CLIP, and reduces the VBench semantic score
by $4.19$ points. Our combined adversarial objective improves ImageNet
convergence and SDXL detail and text alignment over either target alone.
Replacing the linear generator loss with the non-saturating loss slows the
reduction in ImageNet FID. As shown in \cref{fig:gap_fid}(b), our
model first records FID below $1.5$ at 209K iterations, compared with 354K
for the non-saturating variant. The linear loss also gives better image FID
and VBench total score in \cref{tab:ablation}. Our linear objective uses the
discriminator's logit gradient directly, without the additional sample
weighting introduced by the non-saturating loss.
Disabling gap reweighting worsens ImageNet FID and all SDXL metrics in
\cref{tab:ablation}, and lowers VBench total by $1.14$ points. Adapting the
teacher weight across noise levels lowers FID at matched iterations in
\cref{fig:gap_fid}(b), while the real-data coefficient stays fixed.

\section{Conclusion}
\label{sec:conclusion}

We presented \method, a reformulation of distribution matching
distillation that jointly learns from teacher-generated samples
and real data without auxiliary score fitting.
Starting from DMD's score-difference update, \method trains the
student through linear losses on two discriminator logits, which
at the discriminator optimum recover the same distribution-matching
gradient. Training requires no online teacher evaluation and uses
one discriminator update per student update.
We further use the real-data head's gap between real and teacher
samples to adapt teacher supervision across noise levels.
Experiments show improvements over the compared few-step methods
and multi-step teachers in image FID and VBench total score. On MiniMax-H3, users prefer our four-step student over \dmdtwo and rCM, particularly in visual and audio quality.
These results support discriminative learning as an effective approach to distribution matching distillation.

\clearpage
\subsection*{AI use statement}

We used large language models to assist with polishing the writing, retrieving
and discovering related literature, and drafting portions of the manuscript.
The authors take responsibility for the final content of this work, including
its claims, citations, and any text produced with the aid of these tools.

\subsection*{Ethics statement}

This work develops efficient distillation methods for image and video
generation. While faster generation can benefit creative applications and
research, it can also lower the cost of producing misleading or harmful
synthetic media. Distilled models may inherit biases and inappropriate
generation behaviors from their teachers and training data. We encourage
responsible use of these methods, with appropriate safeguards against
impersonation, misinformation, and non-consensual content.

\subsection*{Reproducibility statement}

\Cref{sec:method} describes our method, and Appendix~\ref{sec:proof} provides the
theoretical derivation. The experimental setup and implementation details,
including model architectures, training hyperparameters, and evaluation
protocols, are provided in \cref{sec:exp} and Appendix~\ref{sec:impl}. We will publicly release
our training and evaluation code, dataset preparation and preprocessing
scripts, and experiment configurations to facilitate independent reproduction
and further research.



\bibliography{iclr2027_conference}
\bibliographystyle{plainnat}

\appendix
\section{Proof of \cref{prop:dmd-gan}}
\label{sec:proof}

\paragraph{Optimal discriminator.}
Fix $t$ and $\bm c$ and write $P=p_t$, $Q=p_{\theta,t}$ for the target and
student noisy marginals.  As in the main text, densities are conditional on
$\bm c$, which we omit from their arguments.  Assume positive differentiable
densities on common support and an
unrestricted, unregularized discriminator with equal positive and negative
class weights. The balanced BCE objective is
\begin{equation}
    \mathcal B(h_\phi;p)
    =-\mathbb E_{t,\bm c}\left[
      \mathbb E_{\bm x_t\sim p_t}\log D_\phi(\bm x_t,t,\bm c)
      +\mathbb E_{\bm x_t\sim p_{\theta,t}}
        \log\big(1-D_\phi(\bm x_t,t,\bm c)\big)
    \right],
    \label{eq:disc-loss}
\end{equation}
where $D_\phi=S(h_\phi)$ and
$S(u)=(1+e^{-u})^{-1}$. Pointwise, this is
\begin{equation*}
    -\,P(\bm x)\log S(h(\bm x)) - Q(\bm x)\log\big(1-S(h(\bm x))\big).
\end{equation*}
Using $\tfrac{\partial}{\partial h}\log S(h)=1-S(h)$ and
$\tfrac{\partial}{\partial h}\log(1-S(h))=-S(h)$, the stationarity
condition is $-P(1-S(h))+QS(h)=0$, \ie
$S(h^\star)=P/(P+Q)$, which is the familiar optimal discriminator
of~\citet{goodfellow2014generative}.  Hence,
\begin{equation*}
    h^\star(\bm x)
    =\log\frac{S(h^\star)}{1-S(h^\star)}
    =\log\frac{P(\bm x)}{Q(\bm x)}
    =\log p_t(\bm x)-\log p_{\theta,t}(\bm x).
\end{equation*}
Since the integrand is strictly convex in $h$ for $P,Q>0$, this stationary
point is the unique minimizer.  Differentiating in $\bm x$ yields
\cref{eq:logit-score}.

If positives and negatives instead use different fixed sampling priors
$\pi_+(t,\bm c)$ and $\pi_-(t,\bm c)$, the optimal joint-classification
logit additionally contains $\log[\pi_+(t,\bm c)/\pi_-(t,\bm c)]$.
On their common support, this term is constant in $\bm x$, so its input
gradient vanishes.  The conditional log-ratio expression above assumes
matching priors, not samplewise pairing.

\paragraph{Gradient of the distribution-matching objective.}
Write $\bm x_t=\alpha_tG_\theta(\bm z,\bm c)+\sigma_t\bm\epsilon$ and let
$p_{\theta,t}$ be its density.  The target $p$ does not depend on $\theta$.
Assume the densities and reparameterization are
sufficiently regular to interchange differentiation and integration, with
support independent of $\theta$.  Then
\begin{align*}
    \nabla_\theta D_{\mathrm{KL}}(p_{\theta,t}\|p_t)
    &=\nabla_\theta\,
    \mathbb E_{\bm z,\bm\epsilon}
    \Big[\log p_{\theta,t}(\bm x_t)-\log p_t(\bm x_t)\Big]\\
    &=\mathbb E_{\bm z,\bm\epsilon}
    \left[
      \left(\frac{\partial\bm x_t}{\partial\theta}\right)^{\!\!\mathsf T}
      \big(\bm s_\theta(\bm x_t,t,\bm c)-\bm s_p(\bm x_t,t,\bm c)\big)
    \right]\\
    &\quad+
    \mathbb E_{\bm x_t\sim p_{\theta,t}}\left[
      \left.\nabla_\theta\log p_{\theta,t}(\bm x)\right|_{\bm x=\bm x_t}
    \right],
\end{align*}
where the second term collects the dependence of the density on $\theta$ at
fixed argument.  That term vanishes, since
$\mathbb E_{p_{\theta,t}}[\nabla_\theta\log p_{\theta,t}]
=\int\nabla_\theta p_{\theta,t}=\nabla_\theta\!\int p_{\theta,t}=\bm 0$.
Averaging over $t$ and $\bm c$ gives \cref{eq:kl-grad}. Their sampling
distributions are held fixed as $\theta$ varies.

\paragraph{Equivalence.}
Replace $h_\phi$ in \cref{eq:linear-loss} by the optimal logit at the current
student distribution.  Hold this function fixed during the generator update,
while retaining its derivative with respect to $\bm x_t$.  The chain rule gives
\begin{align*}
    \nabla_\theta\mathcal L_G^{\mathrm{lin}}
    &=-\,\mathbb E\left[
      \left(\frac{\partial\bm x_t}{\partial\theta}\right)^{\!\!\mathsf T}
      \nabla_{\bm x_t}h^\star(\bm x_t,t,\bm c)\right]\\
    &=-\,\mathbb E\left[
      \left(\frac{\partial\bm x_t}{\partial\theta}\right)^{\!\!\mathsf T}
      \big(\bm s_p-\bm s_\theta\big)\right]
    =\nabla_\theta\mathcal L_{\mathrm{DM}}(\theta;p).
\end{align*}
This is a local equality of generator gradients under an optimal, fixed
critic.  It does not differentiate through the dependence of the optimal
critic on $\theta$, or assert equality of the two loss values.

\paragraph{Extension to weighted objectives.}
The same equivalence holds when each noise level is assigned a nonnegative weight $w_t$ that is independent of $\theta$. In particular,
\[
\mathbb E_{t,\bm c}
\left[
w_t D_{\mathrm{KL}}(p_{\theta,t}\|p_t)
\right]
\]
has the same generator gradient as
\[
-\mathbb E
\left[
w_t h^\star(\bm x_t,t,\bm c)
\right],
\]
with $h^\star$ held fixed. This extension does not cover DMD's sample-dependent score normalization. It also treats our gap-based weights as fixed during each generator update rather than differentiating through their evolution over training.

\paragraph{Choice of generator loss.}
For any differentiable $\ell$, a fixed critic gives
\begin{equation}
    \nabla_\theta\,\mathbb E\big[w_t\ell(h_\phi)\big]
    =\mathbb E\left[
      w_t\ell'(h_\phi)
      \left(\frac{\partial\bm x_t}{\partial\theta}\right)^{\!\!\mathsf T}
      \nabla_{\bm x_t}h_\phi
    \right].
    \label{eq:generic-loss}
\end{equation}
Relative to the linear loss, the per-sample multiplier is
$-\ell'(h_\phi)$, which is nonnegative for a non-increasing loss.
The linear loss $\ell(h)=-h$ has multiplier one.  The non-saturating loss
$\ell(h)=-\log S(h)$ instead gives $1-D_\phi$, reducing the contribution
of samples the discriminator already accepts.  At an optimal critic, this
reweights the ideal distribution-matching gradient for each sample and can
change the direction of the aggregate update.
We compare the two losses in \cref{sec:ablation}.

\paragraph{Two fixed targets.}
Applying the result separately to teacher samples and real data gives
\begin{equation}
    h^\star_{T}=\log\frac{p_{T,t}}{p_{\theta,t}},
    \qquad
    h^\star_{R}=\log\frac{p_{R,t}}{p_{\theta,t}}.
    \label{eq:two-logits}
\end{equation}
These are the unrestricted optima of the two balanced classification tasks.
Assume both logits are optimal and the coefficients
$\lambda_{T},\lambda_{R}$ are constant and nonnegative.
With gap reweighting disabled, the generator update is the gradient of
\begin{equation*}
    \mathbb E_{t,\bm c}\!\left[
    \lambda_{T}D_{\mathrm{KL}}(p_{\theta,t}\|p_{T,t})
    +\lambda_{R}D_{\mathrm{KL}}(p_{\theta,t}\|p_{R,t})
    \right].
\end{equation*}
Let $\Lambda=\lambda_{T}+\lambda_{R}>0$ and
$\rho=\lambda_{T}/\Lambda$.  At each fixed $t,\bm c$, define
\begin{equation*}
    M_t(\bm x)=Z_t^{-1}p_{T,t}(\bm x)^\rho
    p_{R,t}(\bm x)^{1-\rho},
    \qquad
    Z_t=\int p_{T,t}(\bm x)^\rho
    p_{R,t}(\bm x)^{1-\rho}\,\mathrm d\bm x.
\end{equation*}
For $0<Z_t<\infty$, the weighted KL sum equals
$\Lambda[D_{\mathrm{KL}}(p_{\theta,t}\|M_t)-\log Z_t]$.
Since $Z_t$ is independent of $\theta$, the corresponding gradient is that of
$\Lambda\mathbb E_{t,\bm c}[D_{\mathrm{KL}}(p_{\theta,t}\|M_t)]$.
The geometric mixture is defined separately at each noise level.  Forward
noising does not generally commute with geometric mixing, and the marginals
$p_{\theta,t}$ all arise from a single clean student distribution.  This
identity does not identify a clean geometric mixture as the global optimum,
or apply as a fixed objective when gap reweighting changes the coefficients.

\clearpage
\section{Implementation Details}
\label{sec:impl}

\paragraph{Training and evaluation configuration.}
We use AdamW with $(\beta_1,\beta_2)=(0,0.99)$ and one discriminator update per generator update. All reported models use generator EMA weights. ImageNet and SDXL use an EMA half-life of $10^6$ generated images with ramp-up factor $0.05$. Wan and MiniMax-H3 uses power-function EMA~\citep{karras2024analyzing} with $\sigma_{\mathrm{rel}}=0.1$. For ImageNet-$64\times64$, we train on 8 H100 GPUs with global batch size 384, learning rates $\eta_G=\eta_D=2\times10^{-6}$, $\lambda_{T}=\lambda_{R}=0.003$, and 570K iterations. For SDXL, we train on 8 H100 GPUs with global batch size 64, $\eta_G=\eta_D=5\times10^{-7}$, $\lambda_{T}=\lambda_{R}=0.005$, and 17K iterations. For Wan2.1-T2V-1.3B, we train on 8 H100 GPUs with global batch size 8, $\eta_G=\eta_D=2\times10^{-6}$, $\lambda_{T}=\lambda_{R}=1$, and 17K iterations. For Wan2.1-T2V-14B, we train on 64 H100 GPUs with global batch size 8, $\eta_G=\eta_D=10^{-6}$, $\lambda_{T}=\lambda_{R}=1$, and 19K iterations, using FSDP and context parallelism with groups of 8 GPUs. For evaluation, we follow the protocols of \dmdtwo~\citep{yin2024improved} for ImageNet and SDXL, and rCM~\citep{zheng2026large} for Wan2.1-T2V. For MiniMax-H3, we train on 8 H200 GPUs with global batch size 8 at a resolution of $1344\times768$ for 800 iterations, using $\eta_G=\eta_D=4\times10^{-5}$ and $\lambda_T=\lambda_R=1$. Both the generator and the discriminator backbone use LoRA with rank and alpha set to 128. For evaluation, we follow the protocols of \dmdtwo~\citep{yin2024improved}
for ImageNet and SDXL, rCM~\citep{zheng2026large} for Wan2.1-T2V and AVGen-Bench~\citep{zhou2026avgenbench} for MiniMax-H3. For user study in MiniMax-H3, we conduct a pairwise user study using 387 prompts from VideoGenEval~\citep{yang2025videogen}. For each prompt, we compare the audio-video output of \method with those of \dmdtwo and rCM, yielding 774 pairs. All models use four sampling steps.
Pairs are presented as A/B comparisons with the left/right placement of \method balanced within each baseline. Each pair is evaluated by three annotators on five dimensions: prompt alignment, visual quality, motion quality, audio quality, and overall preference. For each dimension, annotators select A, B, or a tie. This yields 2,322 pairwise evaluations. For each prompt and dimension, we aggregate the three judgments by majority vote, recording a tie if all three judgments differ. Each baseline therefore has 387 prompt-level outcomes per dimension. We report the preference rate for \method as $N_{\mathrm{win}}/(N_{\mathrm{win}}+N_{\mathrm{loss}})$ over these outcomes, excluding ties. For overall preference, the win/tie/loss counts are 204/129/54 against \dmdtwo and 247/95/45 against rCM, with tie rates of $33.3\%$ and $24.5\%$, respectively.

\paragraph{Discriminator architecture.}
The real and teacher heads share a feature backbone.
For ImageNet and SDXL, the diffusion discriminator uses a pretrained U-Net encoder and middle block with two convolutional heads. Each head downsamples the features using $4\times4$ convolutions, group normalization, and SiLU, then predicts a scalar logit. For the projected-discriminator variant on ImageNet, we follow
D2O-F~\citep{zheng2025revisiting} and use frozen ImageNet-pretrained VGG16-BN and EfficientNet-Lite0 backbones with two trainable sets of multi-scale discriminator heads. For Wan, two MLP heads with a SiLU activation operate on final-block
tokens, and their logits are averaged over the video. For MiniMax-H3, final-block features undergo non-affine RMS normalization
before entering two MLP heads, each consisting of two linear layers with a SiLU activation between them. For noise-band reweighting, ImageNet uses six bands over Karras indices 20--500, grouped by
100 indices, with the last containing only index 500 ($\sigma \approx 2.527$), SDXL uses 100-timestep bands, Wan uses 10
equal-probability bands under its discriminator-time sampling distribution, and MiniMax-H3 uses ten equal-width bands in the base time variable over $[0,1]$. We use $\tau=0.5$ for ImageNet and SDXL training and $\tau=2$ for Wan and MiniMax-H3.

\subsection{Training Speed and Memory}
\label{sec:training-cost}

\Cref{tab:training-cost} compares the online training cost of \dmdtwo
and \method on NVIDIA H200 GPUs, with matched batch sizes and input
shapes within each setting. ImageNet and Wan-14B use full-parameter
training, while MiniMax-H3 uses rank-128 LoRA. We measure
CUDA-synchronized update times after warm-up and peak allocated GPU
memory. Time includes one generator update and its associated critic
updates. Following the official \dmdtwo configuration, we use a 5:1
critic-to-generator update ratio for \dmdtwo, compared with 1:1 for
\method. Evaluation and checkpointing are excluded.

\begin{table}[htbp]
\centering
\caption{Online training time per generator update, including its critic
updates, and peak allocated memory per GPU. Lower is better.}
\label{tab:training-cost}
\small
\setlength{\tabcolsep}{5pt}
\begin{tabular}{lc rr rr}
\toprule
& & \multicolumn{2}{c}{Time (s)} & \multicolumn{2}{c}{Memory (GiB/GPU)} \\
\cmidrule(lr){3-4}\cmidrule(lr){5-6}
Setting & H200 GPUs & DMD2 & Ours & DMD2 & Ours \\
\midrule
ImageNet-$64\times64$ & 1 & 2.321 & \textbf{0.496} & 40.34 & \textbf{33.99} \\
Wan2.1-T2V-14B & 4 & 162.9 & \textbf{38.50} & 131.04 & \textbf{124.56} \\
MiniMax-H3-33B & 4 & 818.1 & \textbf{233.3} & 127.23 & \textbf{114.72} \\
\bottomrule
\end{tabular}
\end{table}

\subsection{Toy Gradient Estimation}
\label{sec:toy-gradient-details}

\paragraph{Distributions.}
We compare two estimates of the same correction field using two-dimensional
Gaussian mixtures with known parameters. The target $p$ represents real data,
and $q$ is a fixed student distribution. For $k=0,\ldots,7$, define
\begin{equation*}
    \bm\mu_k=2\big(\cos(2\pi k/8),\sin(2\pi k/8)\big)^{\mathsf T},
    \qquad \bm\delta=(0.08,-0.04)^{\mathsf T}.
\end{equation*}
The clean distributions are
\begin{align*}
    p(\bm x)&=\sum_{k=0}^7\tfrac18\,
    \mathcal N(\bm x\mid\bm\mu_k,0.35^2\bm I),\\
    q(\bm x)&=\sum_{k=0}^7\pi_k\,
    \mathcal N(\bm x\mid\bm\mu_k+\bm\delta,0.35^2\bm I),
\end{align*}
where $(\pi_0,\ldots,\pi_7)=(0.16,0.14,0.12,0.10,0.10,0.12,0.13,0.13)$.
Adding independent noise $\sigma\bm\epsilon$, with
$\bm\epsilon\sim\mathcal N(\bm0,\bm I)$, gives marginals $p_\sigma$ and
$q_\sigma$. Their component means and weights are unchanged, and their
component covariance is $v_\sigma\bm I$, where $v_\sigma=0.35^2+\sigma^2$.

\paragraph{Analytic reference.}
For a mixture $f_\sigma$ with weights $a_k$ and means $\bm m_k$, its posterior
component weights and score are
\begin{align*}
    r_k(\bm x,\sigma)
    &=\frac{a_k\exp\big(-\|\bm x-\bm m_k\|^2/(2v_\sigma)\big)}
    {\sum_j a_j\exp\big(-\|\bm x-\bm m_j\|^2/(2v_\sigma)\big)},\\
    \bm s(\bm x,\sigma)
    &=\nabla_{\bm x}\log f_\sigma(\bm x)
    =\frac{\sum_k r_k(\bm x,\sigma)\bm m_k-\bm x}{v_\sigma}.
\end{align*}
Applying this expression separately to the target and student gives the
ground-truth field
\begin{equation*}
    \bm g^\star=\bm s_p-\bm s_q
    =\nabla_{\bm x}\log\frac{p_\sigma}{q_\sigma}.
\end{equation*}
Balanced binary classification has the ideal logit
$h^\star=\log(p_\sigma/q_\sigma)$, motivating the comparison with
$\nabla_{\bm x}h_\phi$. Analytic scores and density ratios are used only for
evaluation. Neither estimator receives them as training supervision.

\paragraph{Training.}
We first train a teacher noise predictor on samples drawn directly from $p$
for 20,000 updates. It has two width-64 tanh hidden layers and a linear
two-dimensional output. Its input is
$(x_1/3,x_2/3,\log(\sigma/0.5))$, and its denoising objective is
$\tfrac12\mathbb E\|\epsilon_\omega(\bm x_0+\sigma\bm\epsilon,\sigma)
-\bm\epsilon\|^2$. Noise levels are sampled uniformly from
$\{0.2,0.5,1.0\}$ for each example during pretraining and adaptation.

Both estimators inherit this pretrained teacher. The fake noise predictor
copies all its parameters. The discriminator copies both hidden layers and
uses a new scalar linear head, whose weights have independent Gaussian
initialization with standard deviation $0.02$ and whose bias is zero.
The teacher then stays frozen. All parameters of the fake predictor and
discriminator, including their hidden layers, are adapted for 1,000 updates.
The fake predictor uses denoising on $q$. The discriminator uses balanced
binary cross-entropy on noisy target positives and student negatives.

Pretraining and adaptation use Adam with learning rate $0.001$,
$(\beta_1,\beta_2)=(0.9,0.999)$, and numerical constant $10^{-8}$.
Each adaptation optimizer starts with fresh state. The denoising batch
contains 512 student examples. The classification batch contains 256 target
and 256 student examples, sharing those student inputs with the first half
of the denoising batch. Pretraining also uses batch size 512. Both branches therefore receive 512 training inputs per update.
Computation uses float64.

\paragraph{Evaluation.}
The experiment in \cref{fig:toy-gradient} uses seed 0 and $\sigma=0.5$. The labels
\emph{DMD gradient} and \emph{Ours gradient} refer respectively to
\begin{equation*}
    \widehat{\bm g}_{\mathrm{DMD}}
    =-\frac{\epsilon_{\mathrm{teacher}}}{\sigma}
     +\frac{\epsilon_{\mathrm{fake}}}{\sigma},
    \qquad
    \widehat{\bm g}_{\mathrm{ours}}=\nabla_{\bm x}h_\phi.
\end{equation*}
Both are evaluated on the same $N=2{,}048$ independent held-out samples
$\bm x_i\sim q_{0.5}$ using
\begin{equation*}
    \operatorname{NMSE}(\widehat{\bm g})=
    \frac{N^{-1}\sum_i\|\widehat{\bm g}(\bm x_i)-\bm g^\star(\bm x_i)\|^2}
    {N^{-1}\sum_i\|\bm g^\star(\bm x_i)\|^2}.
\end{equation*}
Their NMSE values are $1.539462$ and $0.144375$. This experiment isolates single-target gradient estimation
with a fixed student distribution. Target samples for both teacher
pretraining and discriminator training are drawn directly from $p$.

\section{Additional Audio-Video Evaluation}
\label{sec:avgenbench}

We follow the evaluation metrics and score aggregation of
AVGen-Bench~\citep{zhou2026avgenbench}, using Qwen3-Omni-Thinking as the
VLM judge. The same H3-enhanced prompts are used for generation and
judging across all methods.
\Cref{tab:h3_avgenbench_thinking} reports all eleven metrics for
MiniMax-H3. \method achieves a total score of $65.4$, compared with
$62.6$ for \dmdtwo and $59.8$ for rCM, while the teacher scores $67.2$.

\begin{table}[htbp]
\centering
\fontencoding{OT1}\selectfont
\caption{\textbf{AVGen-Bench results on MiniMax-H3.} All scores are
normalized to 0--100 ($\uparrow$). Bold marks the best total among distilled models.}
\label{tab:h3_avgenbench_thinking}
\small
\setlength{\tabcolsep}{4pt}
\renewcommand{\arraystretch}{1.22}
\resizebox{\linewidth}{!}{%
\begin{tabular}{lrrrrrrrrrrrr}
\toprule
& & \multicolumn{2}{c}{Quality} & \multicolumn{2}{c}{Synchronization}
& \multicolumn{7}{c}{Fine-grained evaluation} \\
\cmidrule(lr){3-4}\cmidrule(lr){5-6}\cmidrule(lr){7-13}
Method & Total & Visual & Audio & AV & Lip & Face & Lo-Phy & Hi-Phy & Holistic & Text & Speech & Music \\
\midrule
DMD2 & 62.6 & 90.1 & 67.4 & 56.7 & 22.7 & 46.1 & 80.2 & 95.0 & 87.1 & 64.6 & 81.0 & 0.0 \\
rCM & 59.8 & 84.0 & 65.8 & 48.7 & 15.9 & 49.0 & 79.7 & 95.3 & 87.3 & 48.3 & 81.3 & 6.2 \\
\textbf{DMAD (Ours)} & \textbf{65.4} & 93.3 & 70.4 & 49.1 & 30.6 & 53.1 & 82.3 & 93.2 & 87.4 & 70.9 & 80.6 & 11.8 \\
\midrule
Teacher & 67.2 & 92.6 & 70.6 & 58.3 & 39.0 & 59.2 & 79.7 & 94.5 & 87.0 & 74.6 & 84.0 & 1.6 \\
\bottomrule
\end{tabular}}
\end{table}

\section{Additional Qualitative Results}
\label{sec:qualitative}

\subsection{SDXL Comparisons}
\label{sec:sdxl-comparisons}

\begin{figure}[t]
    \centering
    \includegraphics[width=\linewidth]{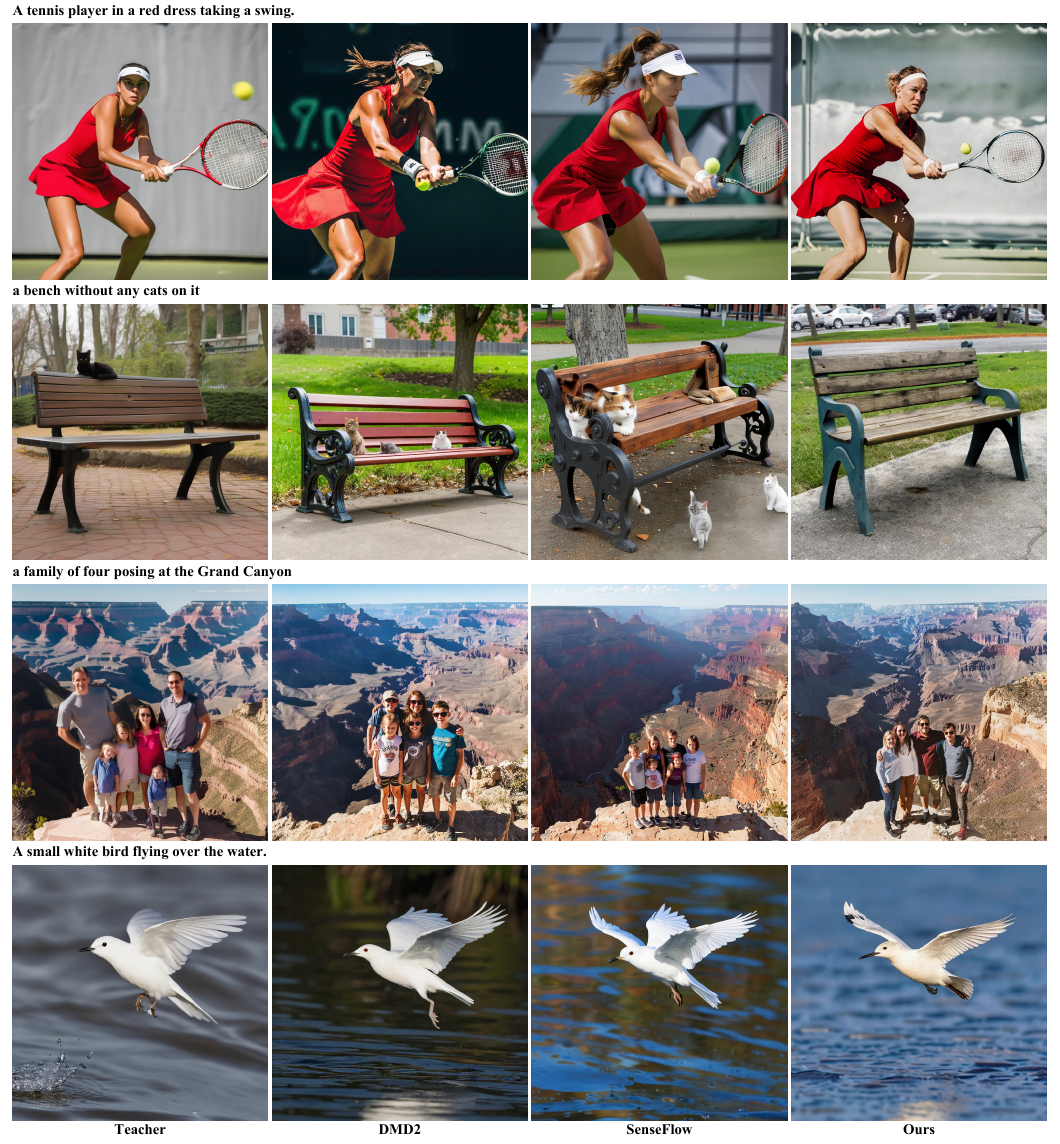}
    \caption{\textbf{Quality comparison of \method with the SDXL teacher,
    \dmdtwo~\citep{yin2024improved}, and SenseFlow~\citep{ge2025senseflow}.}
    All distilled models use 4 sampling steps, while the teacher uses
    50 sampling steps.
    Images in each row share the same text prompt and initial Gaussian noise.
    Our model produces more natural colors and realistic textures with
    better text alignment. (Zoom in for details.)}
    \label{fig:sdxl_comparison}
\end{figure}

Qualitative comparisons in \cref{fig:sdxl_comparison} show that
\method produces more natural colors and textures and follows the
prompts more closely, while some \dmdtwo and SenseFlow samples
appear oversaturated. The tennis example
highlights natural skin texture, while the bird example shows distinct
wing structure and fine feather detail. In the bench example, only our
model satisfies the instruction to exclude cats. It also generates
exactly four people in the family portrait, whereas the teacher,
\dmdtwo, and SenseFlow include extra people.

\subsection{MiniMax-H3 Teaser Prompts}
\label{sec:teaser-prompts}

We list the prompts for the six samples in \cref{fig:teaser}, ordered from left to right and top to bottom.

\begin{enumerate}[leftmargin=1.6em,itemsep=5pt]
    \item A medium shot opens on a seasoned, grey-bearded man in sunglasses and a paisley shirt, Beside him, a younger man in a tank top. In the background, a vibrant mural splashes across a wall, hinting at an urban setting. Faint city murmurs and distant chatter drift in, accompanied by a mellow, soulful hip-hop beat that adds a contemplative yet grounded atmosphere. The older man speaks while the younger man turns toward him, laughs softly, shakes his head in amusement, and then looks back at the camera. The camera slowly pushes in. "The city always got a story," the older man murmurs with a slight nod. The younger man chuckles and replies, "Yeah, you always say that."

    \item A breathtaking, painterly 2D animated continuous visual narrative, rendered with the lush, vibrant, and slightly surreal, almost dreamlike, infused with the intricate, delicate detail of traditional Japanese woodblock prints (Ukiyo-e), follows a young, adventurous, and kind-hearted girl (perhaps with bright, curious eyes and wearing simple, practical, beautifully patterned traditional Japanese farm attire) as she befriends a colossal, gentle, ancient Forest Spirit. The Spirit is a magnificent, awe-inspiring creature, its form a harmonious blend of animal and plant – perhaps with moss-covered, antler-like branches, fur like shimmering leaves that change color with its mood, and eyes like deep, tranquil forest pools. They meet in a sun-dappled, sacred grove deep within an ancient, primeval forest, where impossibly tall, gnarled trees form a living cathedral and tiny, glowing, friendly forest sprites (Kodama-like) peek from behind mossy rocks and giant, fantastical mushrooms. The girl, initially awestruck, offers the massive Spirit a small, carefully cultivated offering – perhaps a perfectly ripe persimmon or a handful of wild berries – her gesture one of pure, innocent respect and affection. The Forest Spirit responds with a slow, gentle inclination of its massive head, its leafy fur rustling like a thousand whispers, and perhaps causes a shower of magical, luminous flower petals to drift down from the canopy, or a tiny, new sapling to sprout at the girl's feet. The animation captures the incredible, detailed textures of the forest, the Spirit's majestic yet gentle presence, and the profound, unspoken emotional connection forming between the child and this ancient guardian of nature. The color palette is a rich symphony of deep forest greens, earthy browns, vibrant floral hues, and the soft, magical glow of the sprites and the Spirit's own subtle luminescence. This continuous, sweeping visual journey is a celebration of the profound, often mystical, bond between humanity and nature, the innocence and courage of childhood, and the power of kindness and respect to bridge even the most fantastical of divides, an affectionate, visually intoxicating ode to ecological harmony and interspecies understanding. The only implied sounds are the gentle rustling of leaves, the distant calls of unseen forest birds, the girl's soft, respectful breathing, the Spirit's deep, resonant, almost inaudible hum, and a soaring, emotionally resonant, orchestral score.

    \item The video opens with a medium, eye-level shot of a beautiful woman with dark hair and warm brown eyes. She wears a magnificent, high-fashion flamingo dress with layers of pink and fuchsia feathers, complemented by whimsical pink, heart-shaped sunglasses. She walks with serene confidence through the crystal-clear, shallow turquoise water of a sun-drenched lagoon. The camera slowly pulls back to a medium-wide shot, revealing the breathtaking scene as the dress's long train glides and floats gracefully on the water's surface behind her. The cinematic, dreamlike atmosphere is enhanced by the vibrant colors of the dress against the serene, minimalist landscape, capturing a moment of pure elegance and high-fashion fantasy.

    \item An adorable happy otter confidently stands on a surfboard wearing a yellow lifejacket, riding along turquoise tropical waters near lush tropical islands, 3D digital render art style.

    \item Static overhead shot of a printed photo of a tree trunk lying on a wooden table.Action Sequence:1, A real human hand enters, places a single pinecone on the paper next to the tree hole, and exits immediately.2, A realistic 3D squirrel emerges from the 2D hole in the photo. The squirrel comes out empty-handed.3, The squirrel sniffs the pinecone sitting on the paper, looks curious, blinks, and tilts its head.4, The squirrel reaches out, grabs that specific pinecone from the table.Key details: Seamless interaction between real world and photo, surreal VFX, squirrel paws are empty initially, heavy weight perception on the pinecone.

    \item An origami dragon unfolds from paper in stop-motion style, then flies over folded mountains. The camera pans with paper dust fluttering midair. Every flap and fold is sharply geometric.
\end{enumerate}

\clearpage
\subsection{Wan2.1-T2V-14B Samples}
\label{sec:wan14b-prompts}

\begin{figure}[htbp]
    \centering
    \includegraphics[width=\linewidth]{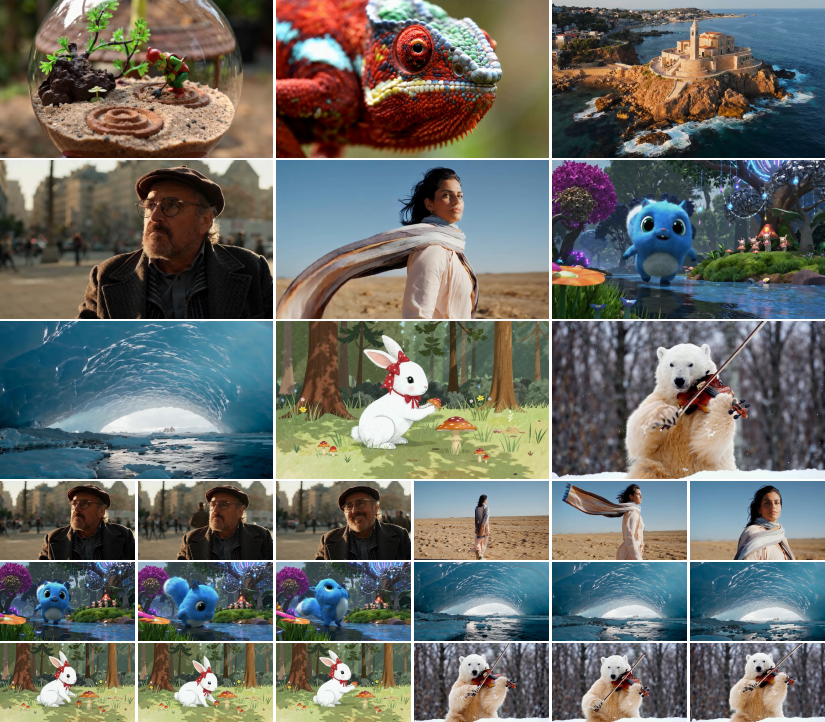}
    \caption{Video samples from our four-step Wan2.1-T2V-14B student. Top: representative frames from nine generated videos.
    Bottom: three frames in chronological order from each of six videos.
    Full prompts are provided in \cref{sec:wan14b-prompts}.}
    \label{fig:wan14b-samples}
\end{figure}

We list the prompts for the nine samples in the upper grid of \cref{fig:wan14b-samples}, ordered from left to right and top to bottom.

\begin{enumerate}[leftmargin=1.6em,itemsep=5pt]
    \item A close up view of a glass sphere that has a zen garden within it. There is a small dwarf in the sphere who is raking the zen garden and creating patterns in the sand.

    \item This close-up shot of a chameleon showcases its striking color changing capabilities. The background is blurred, drawing attention to the animal’s striking appearance.

    \item A drone camera circles around a beautiful historic church built on a rocky outcropping along the Amalfi Coast, the view showcases historic and magnificent architectural details and tiered pathways and patios, waves are seen crashing against the rocks below as the view overlooks the horizon of the coastal waters and hilly landscapes of the Amalfi Coast Italy, several distant people are seen walking and enjoying vistas on patios of the dramatic ocean views, the warm glow of the afternoon sun creates a magical and romantic feeling to the scene, the view is stunning captured with beautiful photography.

    \item An extreme close-up of an gray-haired man with a beard in his 60s, he is deep in thought pondering the history of the universe as he sits at a cafe in Paris, his eyes focus on people offscreen as they walk as he sits mostly motionless, he is dressed in a wool coat suit coat with a button-down shirt , he wears a brown beret and glasses and has a very professorial appearance, and the end he offers a subtle closed-mouth smile as if he found the answer to the mystery of life, the lighting is very cinematic with the golden light and the Parisian streets and city in the background, depth of field, cinematic 35mm film.

    \item A female traveler traverses a desert alone. The camera first shows a distant view of her slightly unsteady footsteps and flowing clothes, then switches to a medium side shot of her scarf billowing in the wind, and finally focuses on her determined gaze.

    \item 3D animation of a small, round, fluffy creature with big, expressive eyes explores a vibrant, enchanted forest. The creature, a whimsical blend of a rabbit and a squirrel, has soft blue fur and a bushy, striped tail. It hops along a sparkling stream, its eyes wide with wonder. The forest is alive with magical elements: flowers that glow and change colors, trees with leaves in shades of purple and silver, and small floating lights that resemble fireflies. The creature stops to interact playfully with a group of tiny, fairy-like beings dancing around a mushroom ring. The creature looks up in awe at a large, glowing tree that seems to be the heart of the forest.

    \item The camera slowly pushes forward into a breathtaking ice cave, its jagged walls sculpted by nature into intricate patterns of blues and whites, reflecting the ethereal light from an opening ahead. The crunch of ice underfoot and the drip-drip of melting water create a serene, echoing soundscape. As the camera moves closer, a gentle, ambient melody begins, swelling with the light from the cave's exit. The camera emerges from the narrow opening into a vast, sun-drenched valley, revealing a group of polar bears playfully sliding down an ice slope, their roars echoing with joy.

    \item In the style of watercolor illustration, a little white rabbit wearing a bow is picking mushrooms in the forest.

    \item A polar bear is playing the violin in the snow.

\end{enumerate}

\clearpage
\subsection{ImageNet-$64{\times}64$ Samples}
\label{sec:imagenet-samples}

\begin{figure}[!ht]
    \centering
    \includegraphics[width=\linewidth]{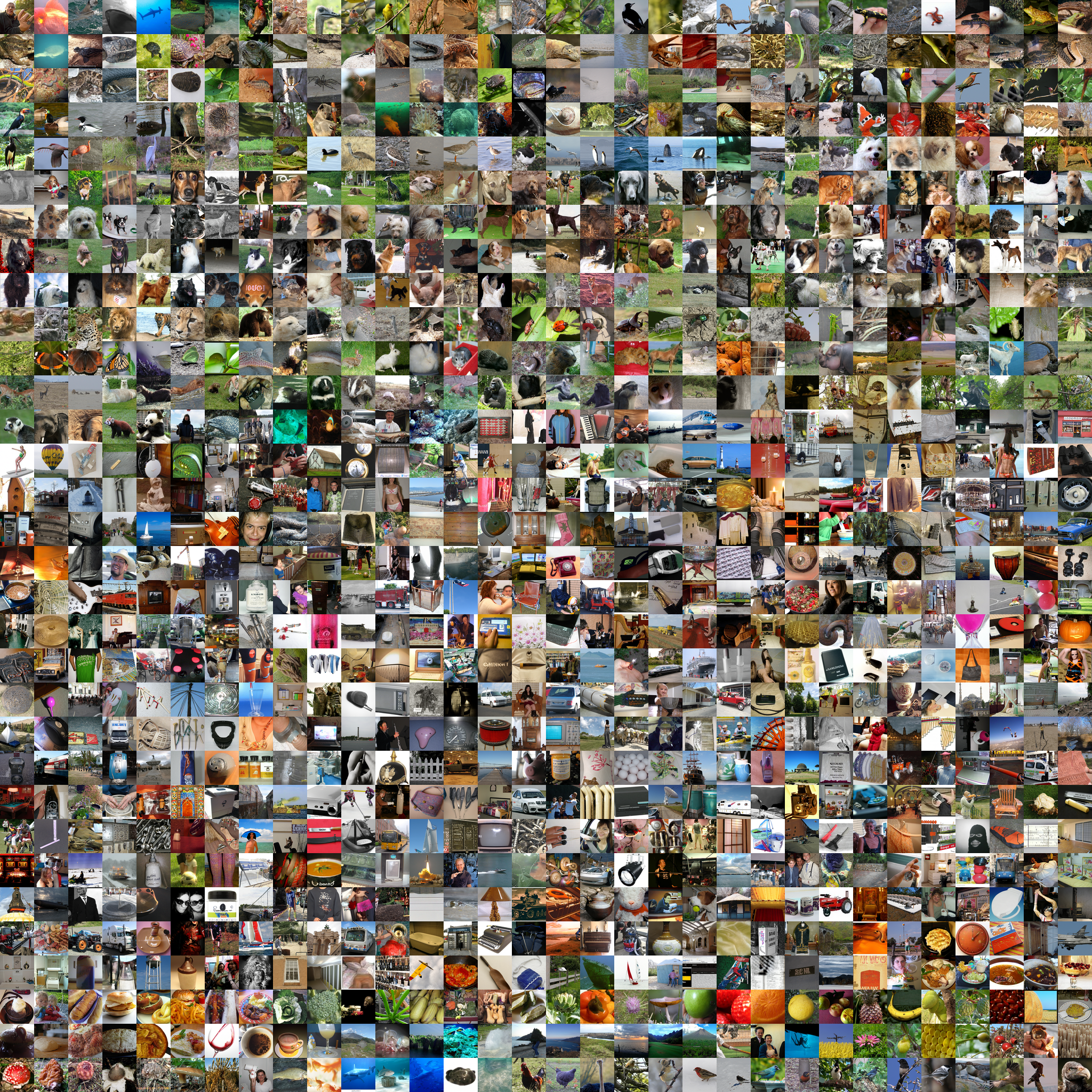}
    \caption{One-step samples from \method on ImageNet-$64{\times}64$ (FID $1.24$).
    Zoom in for details.}
    \label{fig:imagenet-diffusion-samples}
\end{figure}

\begin{figure}[!ht]
    \centering
    \includegraphics[width=\linewidth]{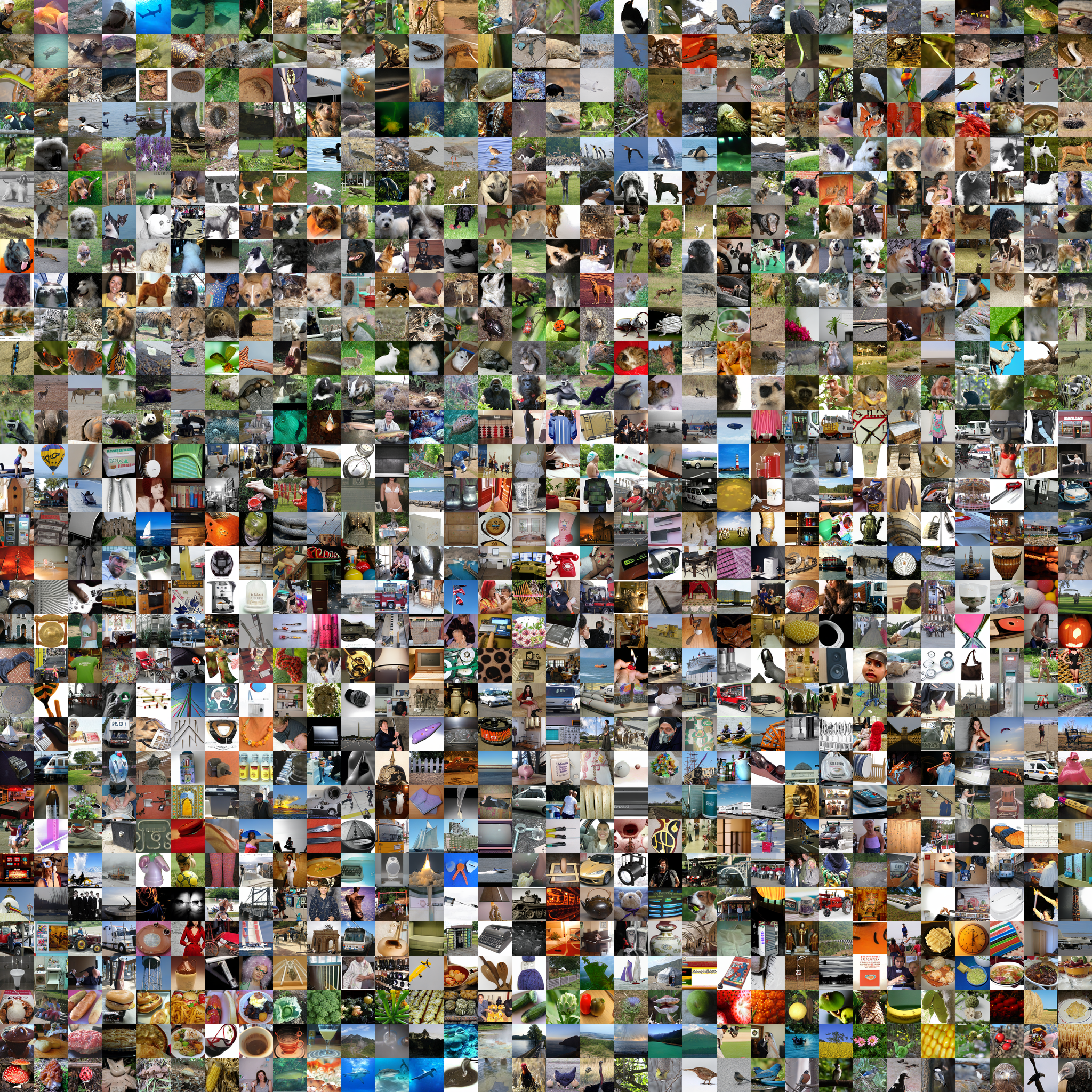}
    \caption{One-step samples from \method on ImageNet-$64{\times}64$ (FID $1.04$).
    Zoom in for details.}
    \label{fig:imagenet-projected-samples}
\end{figure}

\Cref{fig:imagenet-diffusion-samples,fig:imagenet-projected-samples}
show one-step samples from our FID of $1.24$ and FID of $1.04$ results on ImageNet. Each figure shows a $32\times32$ grid of generated images at their native $64\times64$ resolution.

\clearpage

\clearpage
\subsection{More results on MiniMax-H3}
\label{sec:h3-comparisons}

\begin{figure}[!ht]
    \centering
    \includegraphics[width=\linewidth]{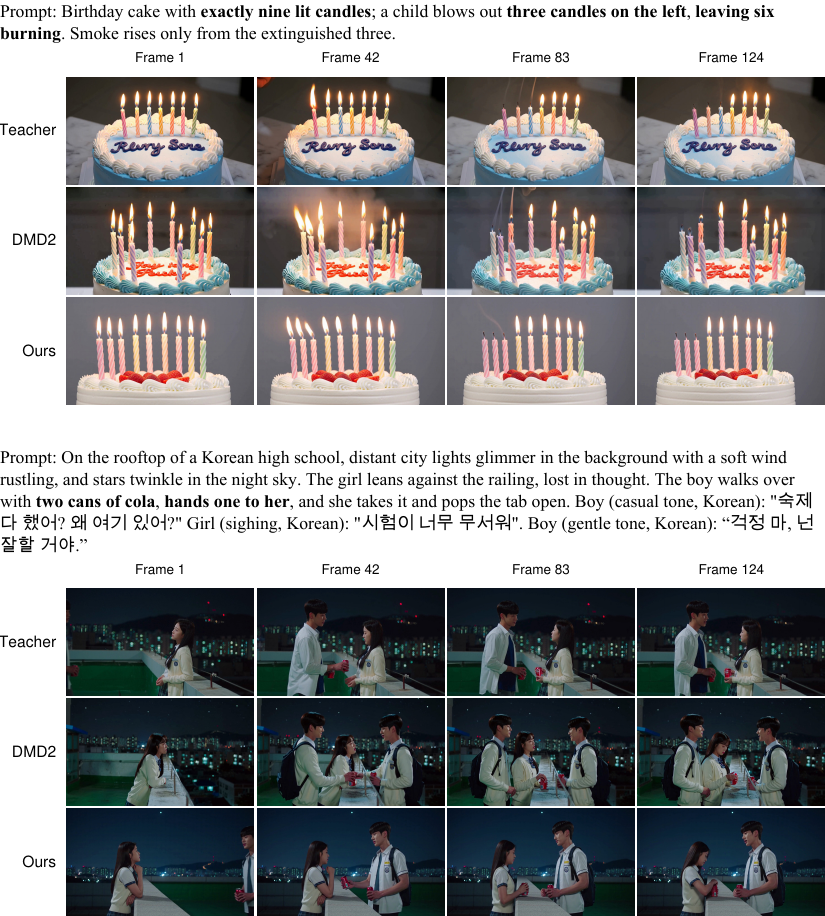}
    \caption{\textbf{Qualitative comparisons of our \method and baseline models on
    MiniMax-H3.} Within each case, rows show Teacher, DMD2, and Ours, from top to bottom.
    Columns show frames 1, 42, 83, and 124, using the same indices for all methods. All compared
    videos are generated using identical initial noise and text prompts.}
    \label{fig:h3-comparison-1}
\end{figure}

\clearpage
\begin{figure}[!ht]
    \centering
    \includegraphics[width=\linewidth]{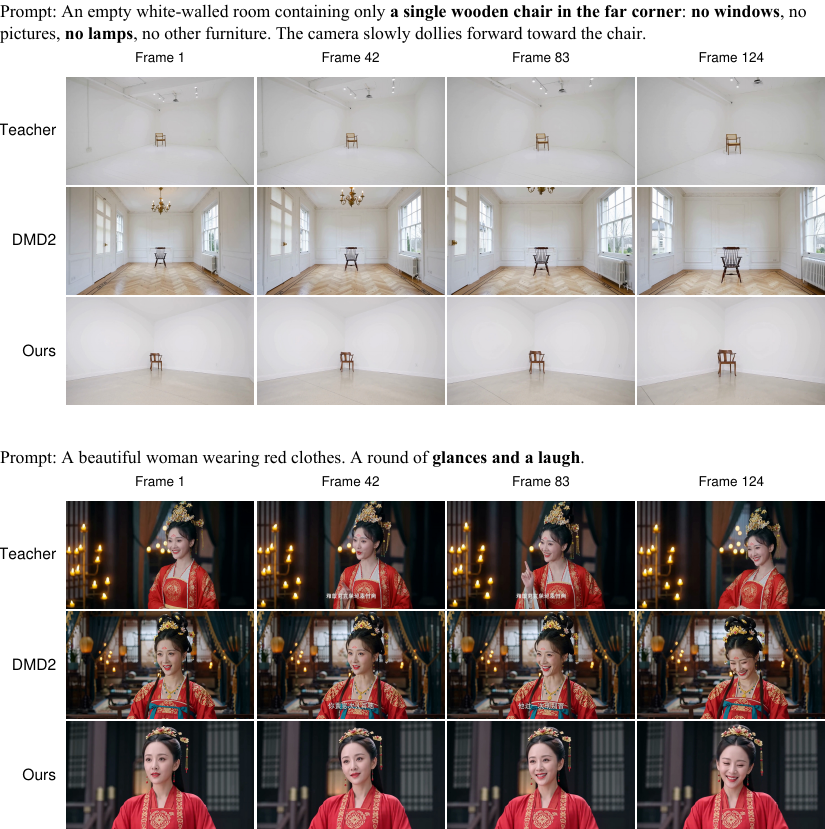}
    \caption{\textbf{Qualitative comparisons of our \method and baseline models on
    MiniMax-H3.} Within each case, rows show Teacher, DMD2, and Ours, from top to bottom.
    Columns show frames 1, 42, 83, and 124, using the same indices for all methods. All compared
    videos are generated using identical initial noise and text prompts.}
    \label{fig:h3-comparison-2}
\end{figure}

\clearpage
\begin{figure}[!ht]
    \centering
    \includegraphics[width=\linewidth]{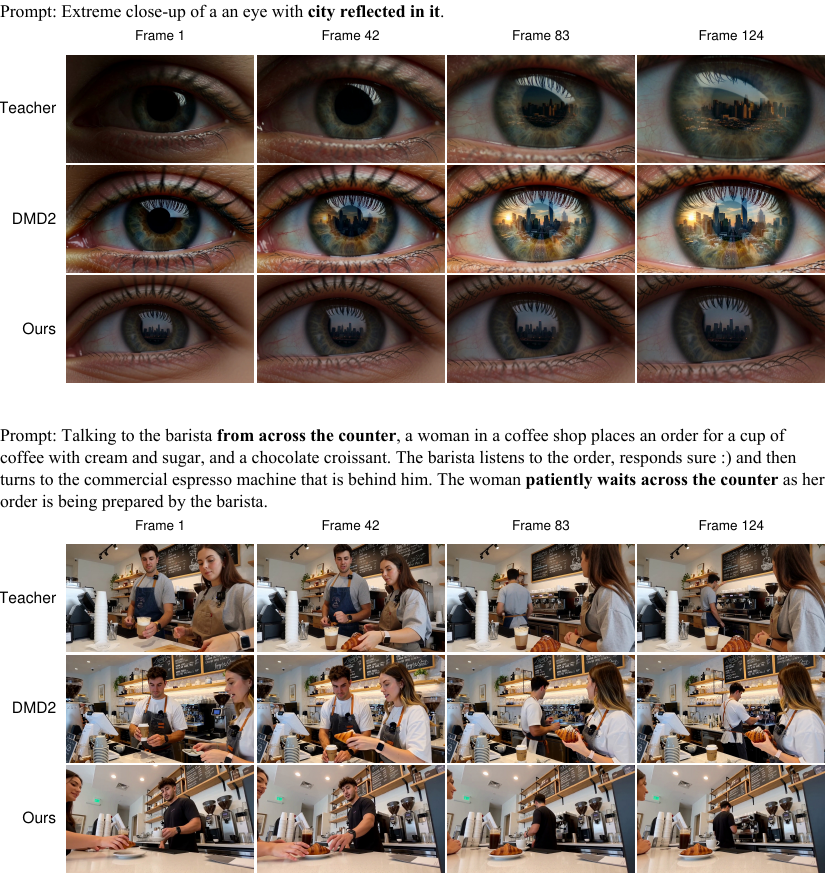}
    \caption{\textbf{Qualitative comparisons of our \method and baseline models on
    MiniMax-H3.} Within each case, rows show Teacher, DMD2, and Ours, from top to bottom.
    Columns show frames 1, 42, 83, and 124, using the same indices for all methods. All compared
    videos are generated using identical initial noise and text prompts.}
\end{figure}

\clearpage
\begin{figure}[!ht]
    \centering
    \includegraphics[width=\linewidth]{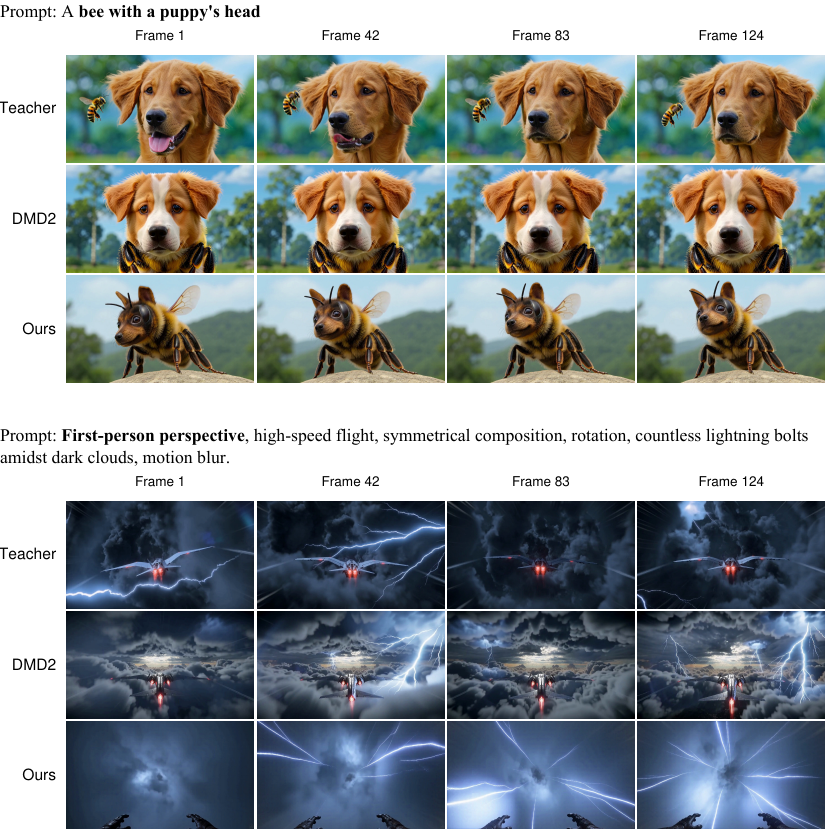}
    \caption{\textbf{Qualitative comparisons of our \method and baseline models on
    MiniMax-H3.} Within each case, rows show Teacher, DMD2, and Ours, from top to bottom.
    Columns show frames 1, 42, 83, and 124, using the same indices for all methods. All compared
    videos are generated using identical initial noise and text prompts.}
\end{figure}

\end{document}